\documentclass{article} 
\usepackage{iclr2026_conference,times}

\usepackage{hyperref}
\usepackage{url}
\usepackage{graphicx} 
\usepackage{natbib}
\usepackage{caption}
\usepackage{booktabs}
\usepackage{amsmath}
\usepackage{amssymb}
\usepackage{xcolor}
\usepackage{tikz}
\usetikzlibrary{arrows.meta, positioning}

\title{\textsc{AgentMercury}: Your Agent Can Synthesize Verifiable Environments for Business Scenarios at scale}

\iclrfinalcopy

\author{Minbyul Jeong \\
Meridian Intelligence Global Inc.\\
wjdalsquf@gmail.com \\
\And
Chanwoong Yoon \\
University of Massachusetts Amherst \\
cwyoon99@gmail.com \\
}
\begin{document}

\maketitle

{
\centering
\href{https://huggingface.co/collections/Minbyul/agentmercury}{%
    \includegraphics[height=1.0em]
    {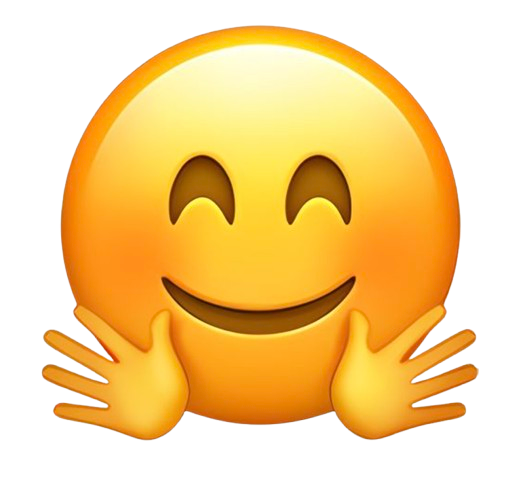}%
    \textbf{Huggingface}
}
\href{https://minstar.github.io/AgentMercury/}{
    \includegraphics[height=1.0em]
    {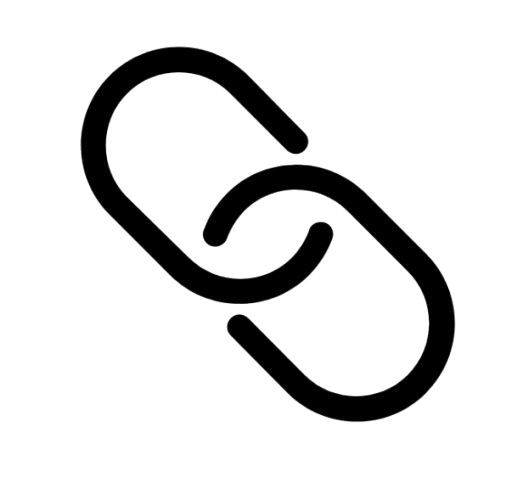}
    \textbf{Project Page}
}
\par
}

\begin{abstract}
Agents learn to act through interaction with environments, yet the environments used for training are often manually constructed or synthesized around predefined tasks and benchmarks.
This task-centric paradigm makes it difficult to scale environments that reflect realistic and evolving workflows where diverse tasks can naturally emerge from the underlying world.
We introduce \textsc{AgentMercury}, a scalable framework for synthesizing executable environments from high-level business scenarios.
Rather than constructing an environment for a specific task, \textsc{AgentMercury} first instantiates a persistent world with entities, services, tools, state, and executable cross-service invariants, from which diverse tasks and interaction trajectories can subsequently emerge.
We construct 4,783 executable environments spanning 14 industries and 50 countries, and use them as training substrates for reinforcement learning.
Despite being generated without targeting the evaluation benchmarks, policies trained on these business-oriented environments improve substantially on both enterprise workflows and out-of-domain benchmarks spanning reasoning, coding, scientific computing, and tool use.
In our experiments, Qwen3.5-4B improves from 12.3 to 15.7 on EnterpriseOps-GYM and from 45.9 to 56.0 on AIME26 after training on \textsc{AgentMercury} environments.
We further show that the construction process itself can be learned: fine-tuning Qwen3.5-35B-A3B on construction traces increases executable-world authoring success from 3.3\% to 83.3\% on held-out business scenarios.
These results show that scenario-grounded environments can provide useful and generalizable learning signals beyond benchmark-specific training, while their construction can itself become a learnable capability.
We release the synthesized environments and tasks, construction code, and trained policy models to support further research on scalable environment generation and agent learning.\footnote{Corresponding Author: Minbyul Jeong}
\end{abstract}

\section{Introduction}
Modern agentic systems learn to interact with environments through complementary capabilities for acting and modeling the world~\citep{nakano2021webgpt}.
A \textit{policy model} learns how to select actions given the agent's current state, while a \textit{world model} predicts how the environment evolves in response to those actions~\citep{sutton1998reinforcement, ha2018world, zuo2026qwen}.
Together, these capabilities define the interaction loop through which an agent acts, observes, and reasons about its environment~\citep{react}.
Yet this loop typically assumes that the world already exists: the entities, services, tools, state, and transition structure of the environment are specified before the agent begins learning.
We argue that this leaves a third role largely outside the learning loop: \emph{constructing the world itself}.
As illustrated in Figure~\ref{fig:hero}, we distinguish three roles:
(1) a \textsc{Planet} role that determines what world exists,
(2) a policy $\pi$ that determines what the agent does within that world, and
(3) a world model $W$ that models how the world responds.
While recent work has substantially advanced policy learning and world modeling, the environment is still generally treated as a predefined artifact whose entities, services, initial state, and transition structure are specified before agent learning begins~\citep{song2026envscaler, xu2026envfactory, wang2026agent, dong2026agent, zuo2026qwen}.
This leaves a complementary question largely outside the learning loop: \emph{how can we systematically construct executable worlds in which agents learn and operate?}

Environments are rarely constructed as standalone worlds; instead, they are typically designed around the tasks that agents are expected to solve.
Manually constructed environments specify entities, tools, states, and transition logic for a particular domain or task~\citep{alfworld, zhou2024webarena, xie2024osworld,
drouin2024workarena}, while synthetic environments are often generated from task descriptions, user instructions, or evaluation specifications~\citep{liu2024agentbench, dong2026agent, zuo2026qwen, team2026kimi}.
This task-centric paradigm has enabled researchers to efficiently build controlled environments for training and evaluating agents, and has driven the rapid development of increasingly capable benchmarks~\citep{jimenez2024swe, toolathlon, mavali2026no}.
However, it also couples the construction of a world to the tasks used to define or evaluate it. 
The resulting environment is therefore often optimized to make a particular task executable or measurable, rather than to represent a broader scenario from which diverse tasks can naturally emerge~\citep{cobbe2020leveraging}.
Consequently, increasing the number of tasks does not necessarily increase the diversity of the underlying worlds~\citep{wang2026agent}, and expanding benchmark coverage does not directly provide a scalable space of interactive environments with richer state and behavior~\citep{dong2026agent}.

Real-world agent applications~\citep{webvoyager, yao2022webshop} instead require environments that capture the complexity of ongoing workflows, such as those represented by SWE~\citep{deng2025swe} and Tau-style domains~\citep{tau2bench, shi2026tau}.
In such settings, a task is rarely an isolated objective defined independently of its surrounding context.
Rather, it emerges from an evolving workflow involving user intents, persistent state, software services, and interactions across multiple systems~\citep{toolllm, apibank, bandi2026mcp, wu2026mcpmark, zheng2026bench}.
A sufficiently rich world can therefore support many distinct tasks and
interaction trajectories without requiring each task to be explicitly specified
during world construction~\citep{terminalworld,phoneworld}.
This motivates a shift from \emph{task-centric environment construction} toward
\emph{scenario-grounded world generation}~\citep{zeng2026glm, team2026kimi},
where a high-level scenario serves as the source from which an executable world
and its possible tasks can emerge.
Realizing this paradigm, however, requires more than generating task descriptions
or static initial states.
The generated world must faithfully instantiate the entities, services, tools,
state representation, transition logic, and world-level invariants that govern
the scenario.
Business workflows provide a natural setting for this formulation because they
combine structured operational processes with diverse interactions among users,
software systems, tools, and evolving persistent state.

\begin{figure*}[t]\centering
\includegraphics[width=\textwidth]{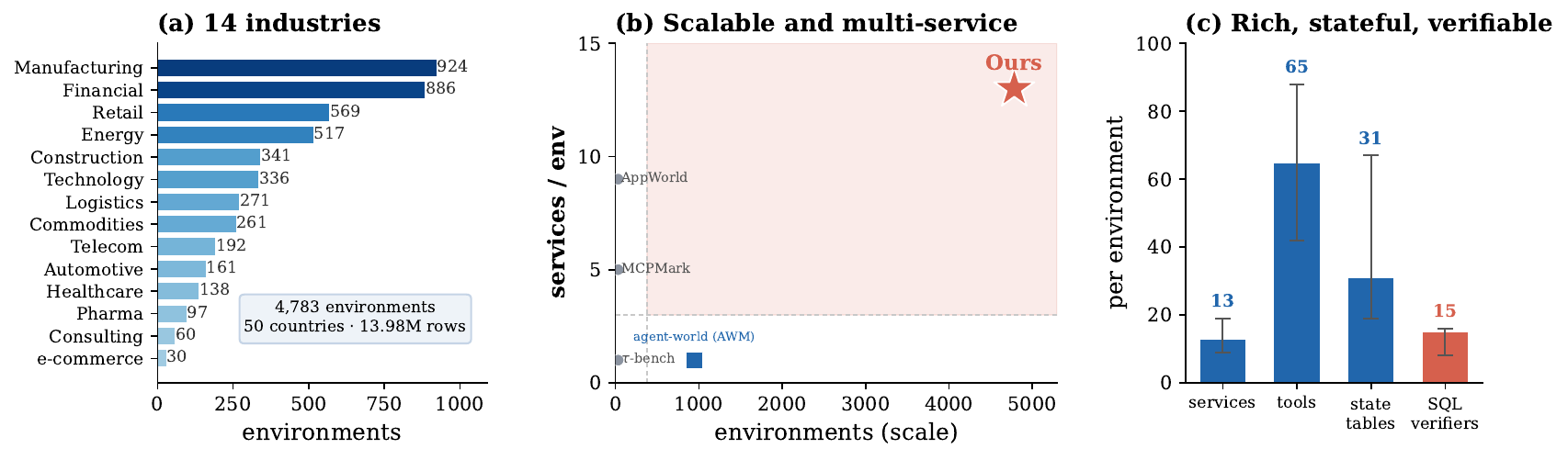}
\caption{
\textbf{\textsc{AgentMercury} synthesizes diverse, scalable, and verifiable business environments at scale.}
\textbf{(a)}~The synthesized environment spans 14 industries and 50 countries, covering 4,783 environments with anonymized to prevent hallucination.
\textbf{(b)}~Comparison of environment scale and multi-service depth with existing agent environments~\citep{appworld, wu2026mcpmark, shi2026tau, dong2026agent}; \textsc{AgentMercury} occupies the regime that is both large-scale and multi-service.
\textbf{(c)}~Each environment contains rich, stateful structure, including multiple services, tools, state tables, and deterministic SQL-based verifiers that enforce cross-service constraints.
}
\label{fig:diversity}
\vspace{-0.2cm}
\end{figure*}

Building on this view, we introduce \textsc{AgentMercury}, a scalable framework for authoring executable environments from high-level business scenarios.
Rather than starting from individual tasks, \textsc{AgentMercury} treats a
scenario as the specification from which an executable world is constructed.
A \textsc{Planet} first instantiates the world, including its entities, services, tools, persistent state, transition dynamics, and world-level invariants.
The resulting world can then give rise to diverse tasks and trajectories through the same underlying workflow, while maintaining consistency across stateful and cross-service interactions.
In particular, the world-level invariants are rendered as executable verification conditions rather than being hard-coded as transition rules, allowing the same world to support different tasks while providing deterministic signals for evaluating their outcomes.
This separation between world construction, task instantiation, and agent interaction enables environments to be generated independently of any particular benchmark task and subsequently reused across training and evaluation.


The resulting environment collection demonstrates that scenario-grounded construction can scale beyond isolated task instances.
As shown in Figure~\ref{fig:diversity}, \textsc{AgentMercury} synthesizes 4,783 executable environments spanning 14 industries and 50 countries.
Compared to previous works~\citep{appworld, wu2026mcpmark, shi2026tau, dong2026agent}, our synthesized environments and tasks are competitive in scale while providing substantially richer multi-service and stateful interactions.
Each environment contains persistent state, multiple services and tools, and executable cross-service constraints that can be verified through interaction.
Rather than defining an environment around a single task, this construction process creates a persistent world from which diverse tasks and interaction trajectories can emerge. The resulting collection therefore provides a broad substrate for training agents in worlds that are independent of the target evaluation tasks.

We first investigate whether these scenario-grounded environments can serve as effective training substrates for policy learning.
We train policy models with reinforcement learning directly in environments synthesized by \textsc{AgentMercury}, without constructing the training environments around the target benchmark tasks.
Within these worlds, policies must interact with persistent state, use available tools, and satisfy constraints induced by the underlying business scenarios.
As training progresses, the resulting policies improve their rewards on these executable tasks and, importantly, also improve on established benchmarks that were not used to construct the training environments.
We observe gains across enterprise workflows as well as out-of-domain benchmarks spanning reasoning, coding, scientific computing, knowledge, and tool use.
These results suggest that scaling the diversity of scenario-grounded worlds can provide learning signals that extend beyond the specific tasks used during training.

We next ask whether the construction process itself can become a learnable capability.
\textsc{AgentMercury} exposes the intermediate construction traces used to transform high-level business scenarios into executable worlds, including changes to entities, services, persistent state, transition logic, and verification conditions.
These traces provide structured supervision for learning how executable worlds are constructed and modified.
Using this supervision, we fine-tune a policy model to author executable worlds from previously unseen business scenarios.
The resulting model substantially improves its success rate on held-out authoring tasks, demonstrating that environment construction is not only an engineering process performed outside the learning loop, but can itself be learned from the construction traces generated by \textsc{AgentMercury}.
This points toward a scalable loop in which scenario-grounded worlds provide training signals for agents, while the construction process provides a learnable interface for expanding the space of available worlds.

In summary, our contributions are as follows:
(1) We introduce \textsc{AgentMercury}, a scalable framework for scenario-grounded synthesis of executable environments, making the construction of the agent's world an explicit \textsc{Planet} role in the agent-environment system.
(2) We construct and release 4,783 executable business environments spanning 14 industries and 50 countries, with persistent state, multi-service interactions, and executable cross-service constraints, providing a reusable substrate for reinforcement learning.
(3) We demonstrate that policies trained in these environments improve both in-domain enterprise workflows and a broad set of out-of-domain benchmarks covering reasoning, coding, scientific computing, knowledge, and tool use, despite the training environments being constructed independently of the target evaluation tasks.
(4) We show that the construction traces generated by \textsc{AgentMercury} can themselves provide effective supervision for environment authoring. Fine-tuning on these traces enables models to construct executable worlds from high-level business scenarios, suggesting that world construction can become a learnable capability.
(5) We release the synthesized environments, business scenarios and task instructions, construction code differences, and trained policy models to facilitate further research on scalable environment generation and agent learning.

\section{Preliminaries}
\label{sec:preliminaries}

\paragraph{Policy Models.}
A policy model specifies how an agent acts within an environment.
At each timestep $t$, a policy $\pi$ maps the agent's available information to a distribution over actions:
\begin{equation}
    a_t \sim \pi(\cdot \mid h_t),
\end{equation}
where $h_t$ denotes the interaction history available to the agent up to timestep $t$.
In a fully observable setting, $h_t$ may reduce to the current state $s_t$; in the partially observable environments considered here, however, the underlying state is not directly exposed to the agent.
The policy therefore determines \emph{how the agent acts} within a given world, but does not determine what entities, services, or transition mechanisms constitute that world~\citep{sutton1998reinforcement}.

\paragraph{World Models.}
A world model provides a predictive representation of how an environment evolves in response to agent actions~\citep{ha2018world, hafner2019dream, hafner2023mastering}.
Given an interaction history and an action, a world model predicts the subsequent state or observation:
\begin{equation}
    W(s_{t+1} \mid h_t, a_t).
\end{equation}
Thus, while the environment executes its transition dynamics, a world model learns to approximate those dynamics and can be used for prediction, planning, or simulation~\citep{schrittwieser2020mastering}.
Importantly, the world model operates on an environment that is already defined: it models \emph{how an existing world evolves}, rather than determining \emph{what world exists}.

\paragraph{Executable Environments.}
We consider an environment as an executable representation of a business scenario in which an agent can directly interact with software services~\citep{zhou2024webarena, drouin2024workarena}.
At timestep $t$, the environment maintains a state $s_t$ consisting of the current state of the underlying software, including persistent data and service-level state~\citep{taubench, xie2024osworld}.
The agent interacts with the environment through executable actions, such as tool calls to available services, and receives observations $\omega_{t+1}$ as the resulting tool outputs:
\begin{equation}
    a_t \sim \pi(\cdot \mid s_t),
    \qquad
    s_{t+1} = T(s_t, a_t),
    \qquad
    \omega_{t+1} \sim O(\cdot \mid s_{t+1}, a_t).
\end{equation}
Here, $T$ is the executable transition function implemented by the environment, rather than a model inferred during interaction.
A world model $W$ instead provides a predictive approximation of these dynamics:
\begin{equation}
    W(s_{t+1} \mid s_t, a_t).
\end{equation}

The environment is therefore a concrete executable world that determines what states and actions are possible and how the world responds to them.
In our setting, this world is instantiated by business software, with persistent data, services, tools, and transition logic forming the underlying execution substrate.
Importantly, environment execution presupposes an environment specification: the entities, services, initial state, transition logic, and constraints must be determined before an agent can interact with the resulting world.
Existing environments typically obtain this specification through manual design or synthetic construction~\citep{zhou2024webarena,xie2024osworld,drouin2024workarena,dong2026agent,wang2026agent,zuo2026qwen}.
This separation between constructing a world and acting within a world motivates the additional role introduced in the following section.

\section{Method}
\label{sec:planet}

\subsection{Overview}
\label{sec:method_overview}
\begin{figure}[t]
    \centering
    \includegraphics[width=\columnwidth]{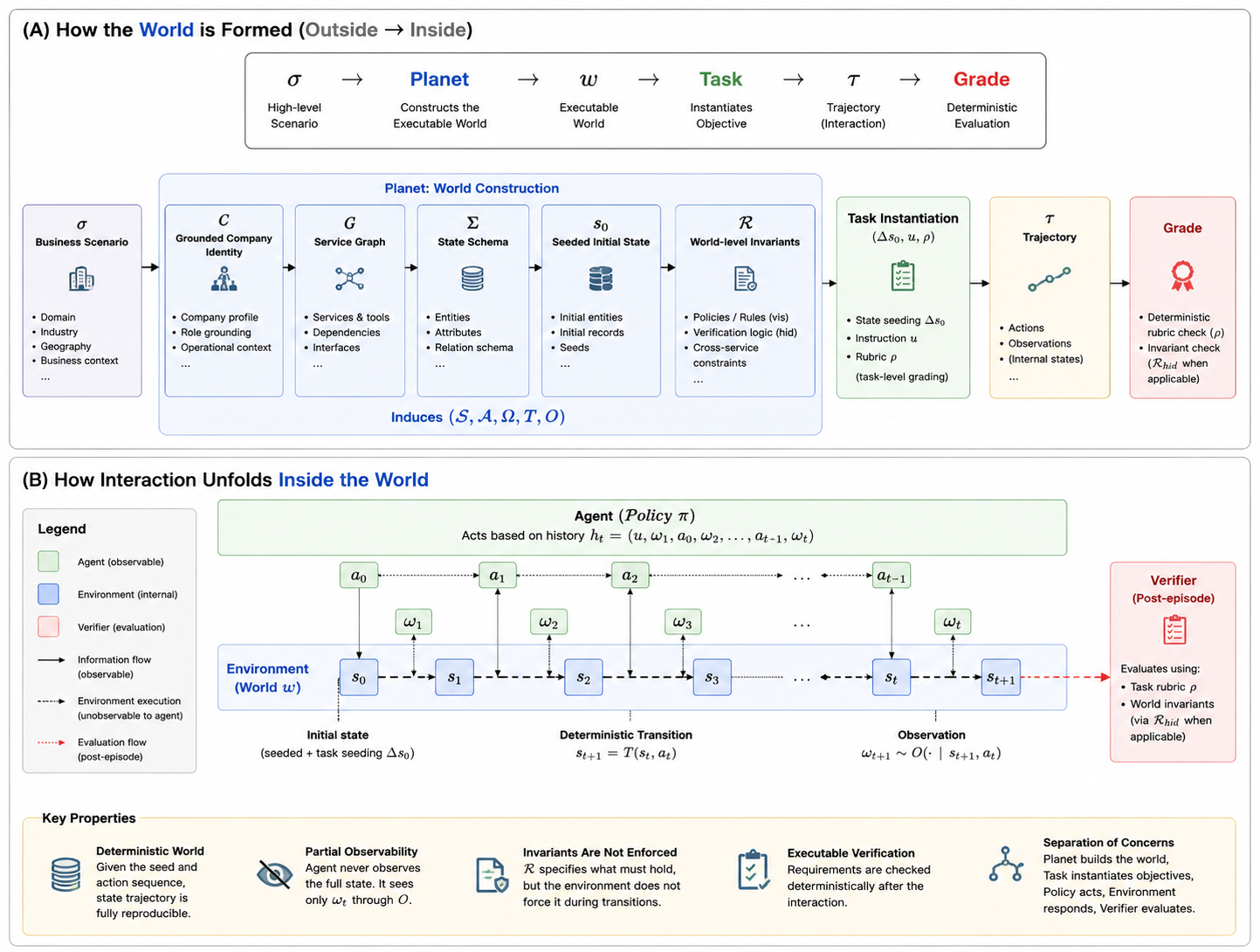}
    \caption{
    \textbf{From scenario-grounded world construction to agent interaction.}
    \textbf{(A)}~\textsc{Planet} transforms a high-level business scenario $\sigma$ into an executable world $w$, constructing its company identity, service graph, state schema, seeded initial state $s_0$, and world-level invariants $\mathcal{R}$. A task is then instantiated from the resulting world through task-specific state seeding and a rubric. \textbf{(B)}~Within the constructed world, a policy acts from the interaction history, producing actions that deterministically transition the environment from $s_t$ to $s_{t+1}$, while observations $\omega_{t+1}$ are exposed through the observation interface. The resulting trajectory is evaluated post-episode using the task rubric and, where applicable, the world-level invariants. The formulation thus separates world construction, task instantiation, interaction, and evaluation.
    }
    \label{fig:hero}
    \vspace{-0.2cm}
\end{figure}

\textsc{AgentMercury} treats environment construction as an explicit role within the agent--environment system.
Given a high-level business scenario $\sigma$, the system first constructs an executable world, from which task instances are subsequently instantiated.
Agents then interact with the resulting world through executable tools, and their trajectories are evaluated against deterministic task- and world-level constraints.
The overall construction and interaction process can be summarized as
\begin{equation}
    \sigma
    \xrightarrow{\textsc{Planet}}
    w
    \xrightarrow{\textsc{Task}}
    (u,\rho)
    \xrightarrow{\pi}
    \tau
    \xrightarrow{\textsc{Grade}}
    r,
\end{equation}
where $w$ denotes an executable world, $u$ is a task instruction,
$\rho$ is its task-specific grading specification, $\tau$ is an interaction trajectory, and $r$ is the resulting reward.
This decomposition separates world construction from task specification: the world is generated once from a scenario, while multiple tasks and trajectories can subsequently emerge from the same underlying world.
We depict this overall flow at the Figure~\ref{fig:hero}(A).

\subsection{Planet: Scenario-to-World Construction}
\label{sec:planet}

Existing agent environments are typically provided as fixed artifacts, constructed manually or through task-conditioned synthesis~\citep{song2026envscaler, xu2026envfactory, xu2026envfactory, song2026envscaler, xu2026theagentcompany}.
In contrast, we introduce \textbf{\textsc{Planet}} as an explicit environment-authoring role that constructs an executable world from a high-level business scenario.
The input to \textsc{Planet} is a scenario $\sigma$ describing the business context, while its output is a complete world specification that can be instantiated and executed by the environment runtime.

We represent an executable world as
\begin{equation}
    w =
    \langle
    \mathcal{S},
    \mathcal{A},
    \Omega,
    T,
    O,
    s_0,
    \mathcal{R}
    \rangle,
    \label{eq:world}
\end{equation}
where $\mathcal{S}$, $\mathcal{A}$, and $\Omega$ denote the state, action, and observation spaces, respectively.
$T$ is the executable transition function, $O$ is the observation function, and $s_0$ is the seeded initial state.
The set $\mathcal{R}$ denotes the world-level invariants: properties that the world requires to hold but does not itself enforce during state transitions.
Each invariant is associated with an executable verification condition, allowing the resulting world to be evaluated deterministically.

The central operation of \textsc{Planet} is therefore
\begin{equation}
    w \sim \textsc{Planet}(\cdot \mid \sigma).
    \label{eq:planet}
\end{equation}
Rather than directly generating an individual task, \textsc{Planet} constructs the underlying world from which many tasks can be instantiated.
The resulting world contains the entities, services, tools, state, executable transition logic, and world-level invariants required to represent the business scenario.

\paragraph{Structured World Synthesis.}
The construction process decomposes scenario grounding into several structured artifacts.
Let $C$ denote the grounded company identity, $G$ the service graph, and $\Sigma$ the state schema.
The world-generation process can be factorized as
\begin{equation}
\begin{split}
\textsc{Planet}(w \mid \sigma)
&=
p(C \mid \sigma)\,
p(G \mid C)\,
p(\Sigma \mid G,C) \\
&\quad \times
p(s_0 \mid \Sigma)\,
p(\mathcal{R} \mid G,\Sigma,s_0),
\end{split}
\label{eq:planet_factorization}
\end{equation}
where $(\mathcal{S},\mathcal{A},\Omega,T)$ are determined by the resulting service graph and state schema.
This decomposition separates the construction of the world structure from the construction of its invariant set.
In particular, $G$ and $\Sigma$ determine the executable software structure and transition dynamics, while $\mathcal{R}$ specifies properties that should hold across the resulting world.

Importantly, $\mathcal{R}$ is not part of the transition mechanism. 
The executable transition function $T$ applies agent actions to the
underlying software state, whereas $\mathcal{R}$ defines conditions that can subsequently be checked against that state.
This distinction allows the environment to represent workflows in which the agent is responsible for satisfying cross-service requirements rather than having those requirements automatically enforced by the simulator.
For example, a cross-service invariant may require a downstream record to exist after an upstream business event, while the environment does not automatically create that record on the agent's behalf.

\paragraph{Visible and Hidden Views of World Invariants.}
The same world-level invariants can appear in different forms during interaction and evaluation.
We therefore distinguish their visible and hidden views as
\begin{equation}
    \mathcal{R}_{\mathrm{vis}}
    =
    \operatorname{view}_{\mathrm{vis}}(\mathcal{R}),
    \qquad
    \mathcal{R}_{\mathrm{hid}}
    =
    \operatorname{view}_{\mathrm{hid}}(\mathcal{R}).
    \label{eq:rule_views}
\end{equation}
The visible view may be surfaced through in-world documents or policies that the agent must discover through interaction, rather than being given as an oracle at the beginning of a task.
The hidden view contains executable verification conditions used for
deterministic evaluation.
Thus, $\mathcal{R}_{\mathrm{vis}}$ and $\mathcal{R}_{\mathrm{hid}}$ are two views of the same underlying invariants rather than independent rule sets.

\subsection{Task Instantiation from a World}
\label{sec:task}
Once an executable world has been constructed, tasks are generated from the resulting world rather than used as the specification from which the world itself is built.
This allows different tasks to share the same underlying services, entities, state schema, and transition dynamics.

A task generator receives the initial world state and its invariants and produces a user objective together with a task-specific grading specification:
\begin{equation}
    (\Delta s_0,u,\rho)
    \sim
    \textsc{Task}(\cdot \mid s_0,\mathcal{R}),
    \label{eq:task}
\end{equation}
where $u$ denotes the task instruction, $\Delta s_0$ denotes task-specific state seeding, and $\rho$ denotes the task rubric.
The seeded task state is applied to the world before interaction begins, yielding the initial state for that particular task.
Consequently, multiple tasks may share the same world structure while starting from different task-specific states.

The task rubric $\rho$ is derived from the world-level structure together with the task-specific state changes.
It specifies the task-level assertions required for successful completion, while $\mathcal{R}$ continues to describe invariants of the underlying world.
This separation prevents task-specific objectives from being conflated with the rules that define the world itself.

\subsection{Executable Agent--Environment Interaction}
\label{sec:interaction}
In Figure~\ref{fig:hero}(B), we describe the overall interaction between agent and executable environment.
After task instantiation, the agent interacts directly with the executable world through its available tools.
At timestep $t$, the environment maintains a state $s_t$, and the agent selects an action according to its policy $a_t \sim \pi(\cdot \mid h_t)$, where $h_t$ denotes the interaction history available to the agent.
The world state itself is not directly exposed; observations enter the history through the environment's observation channel.

The environment then executes the selected action and deterministically updates its underlying state $s_{t+1} = T(s_t,a_t)$.
The resulting state is exposed to the agent only through an observation $\omega_{t+1} \sim O(\cdot \mid s_{t+1},a_t)$, which is appended to the interaction history $h_{t+1} = (h_t,a_t,\omega_{t+1})$.
This yields the interaction loop
\begin{equation}
    s_t
    \xrightarrow{\;\pi\;}
    a_t
    \xrightarrow{\;T\;}
    s_{t+1}
    \xrightarrow{\;O\;}
    \omega_{t+1}.
    \label{eq:interaction_loop}
\end{equation}

Because the transition function is executed by the environment rather than predicted by a learned model, the resulting trajectories are reproducible under a fixed initial seed and action sequence.
This property also enables trajectories to be replayed and deterministically re-evaluated after interaction.

A trajectory of horizon $H$ is given by
\begin{equation}
    \tau =
    (s_0,a_0,s_1,\omega_1,
     a_1,s_2,\omega_2,\ldots,
     a_{H-1},s_H,\omega_H).
    \label{eq:trajectory}
\end{equation}
Although a learned world model $W$, such as Qwen-AgentWorld\footnote{https://huggingface.co/Qwen/Qwen-AgentWorld-35B-A3B}~\citep{zuo2026qwen}, could in principle approximate the transition process,
\begin{equation}
    W(s_{t+1}\mid s_t,a_t,\mathcal{R})
    \approx
    T(s_{t+1}\mid s_t,a_t),
    \label{eq:world_model}
\end{equation}
we do not train a separate world model in this work.
Instead, \textsc{AgentMercury} uses the executable environment itself as
the source of transition dynamics and focuses on the construction of such
worlds and their use as training substrates for policy learning.

\subsection{Deterministic Grading}
\label{sec:grading}

The executable world provides not only an interaction substrate but also
a deterministic basis for evaluating agent behavior.
Given a completed trajectory $\tau$, the grader evaluates the final state
and task-specific assertions against the corresponding rubric:
\begin{equation}
    r(\tau)
    =
    \textsc{Grade}
    (s_H,\tau;\rho,\mathcal{R}_{\mathrm{hid}})
    \in [0,1].
    \label{eq:grade}
\end{equation}
The task rubric $\rho$ captures task-specific success conditions, while
$\mathcal{R}_{\mathrm{hid}}$ provides hidden world-level verification
conditions.
Whenever possible, these conditions are evaluated through deterministic
checks over the underlying database state rather than relying solely on
model-based judgment.

Because the transition dynamics are deterministic and the initial state is seeded, the final state can be reconstructed from the initial state and the executed action sequence.
We refer this as a golden reasoning trace.
This enables the same trajectory to be replayed and re-scored independently of the original interaction process.
The separation between execution and grading is particularly important for cross-service constraints: such constraints need not be enforced during execution, but can instead be checked after the agent completes its workflow.

\subsection{Policy Learning on Synthesized Worlds}
\label{sec:policy_training}
Still, many of researcher are working hard to make agents as a policy model.
We thus provide how our synthesized environments could help working as a policy model.
Unlike task-centric environment construction, the environments used during training are generated from business scenarios independently of the target evaluation tasks.
A policy therefore learns to operate over a distribution of executable worlds containing diverse services, tools, states, workflows, and cross-service constraints.

For a policy parameterized by $\theta$, training optimizes its behavior using rewards obtained from the deterministic environment grader:
\begin{equation}
    a_t
    \sim
    \pi_\theta(\cdot\mid h_t),
    \qquad
    r(\tau)
    =
    \textsc{Grade}
    (s_H,\tau;\rho,\mathcal{R}_{\mathrm{hid}}).
    \label{eq:policy_training}
\end{equation}
The resulting training loop therefore couples policy learning to executable worlds without requiring the environment generator to enumerate the tasks that the policy will encounter.

Importantly, \textbf{\textsc{AgentMercury}} separates the authoring of worlds from the optimization of policies.
\textbf{\textsc{Planet}} determines what world exists, \textbf{Task}
determines what objective is posed within that world, and the policy determines how the agent acts to satisfy that objective.
The executable transition function determines how the world responds, while deterministic grading determines whether the resulting trajectory satisfies the intended requirements.
This separation enables the same synthesized world distribution to support multiple tasks, trajectories, and policy-training configurations.

\section{Experiments}
We evaluate \textsc{AgentMercury} from two complementary perspectives.
First, we ask whether executable worlds synthesized from high-level business scenarios provide effective training signals for agent policy learning:
\textbf{Can synthesized worlds support effective and stable policy optimization on business-oriented agent tasks?}
Second, we ask whether the environment-construction process itself can be learned by an agent:
\textbf{Can agents learn to construct executable worlds from high-level business scenarios?}
The first question evaluates the world as a training substrate, while the second evaluates whether the construction process exposed by \textsc{AgentMercury} can itself become a learnable agent capability.

\subsection{Experimental Setup}
We use Qwen3.5-4B and Qwen3.5-35B-A3B~\citep{qwen3.5} as the primary training models and optimize the policy with
group relative policy optimization (GRPO)~\citep{shao2024deepseekmath} with coefficient details such as Dr. GRPO~\citep{liu2025understanding}.
To demonstrate that our environments are not depend on specific RL algorithm, we also try a single-rollout asynchronous optimization (SAO)~\citep{hou2026single} for our synthesized tasks.
The training corpus contains 43,300 task instances generated from the synthesized environments, with multiple task seeds per environment.
More details about task setup is in Appendix~\ref{app:tasks}.

We train policy models directly in the executable environments synthesized by \textsc{AgentMercury} and evaluate the resulting policies on \textsc{EnterpriseOps-Gym}~\citep{malay2026enterpriseops}, a business-oriented agent benchmark covering multi-service workflows and tool-based interactions.
Also, we evaluate several out-of-domain benchmarks to measure knowledge, reasoning, and tool use abilities such as AIME26~\citep{aime26}, HMMT~\citep{dekoninck2026matharena}, LiveCodeBench v5 and v6~\citep{jain2025livecodebench}, SciCode~\citep{tian2024scicode}, tau-3~\citep{shi2026tau}, BFCL~\citep{patil2025berkeley}, and GPQA-Diamond~\citep{rein2023gpqa}.
For each benchmark, we perform three independent evaluation runs and report the mean and standard deviation across runs.
Our training environments are constructed independently of all the benchmark tasks.
This separation allows us to test whether scenario-grounded environments provide transferable learning signals rather than merely reproducing the target benchmark.
We detailed the overall experimental setup with hyperparameter details in Appendix~\ref{app:hyperparameter} and benchmark details in Appendix~\ref{app:benchmark}.

\subsection{Policy Optimization of Business-oriented Tasks}
We first ask whether training on environments synthesized by
\textsc{AgentMercury} leads to capabilities that transfer beyond the environments encountered during training.
Importantly, our training environments are constructed from business scenarios and are not designed around any of the evaluation benchmarks.
We therefore evaluate the resulting policy on a diverse set of established out-of-domain benchmarks spanning mathematical reasoning, scientific computing, competitive programming, knowledge-intensive reasoning, and interactive tool use.
This evaluation provides a stringent test of whether the learning signals provided by synthesized environments capture general agentic capabilities rather than benchmark-specific behaviors.

\begin{table}[t]
\caption{Result on \textsc{EnterpriseOps-Gym}~\citep{malay2026enterpriseops}.
We evaluate whether training on environments and tasks synthesized by \textsc{AgentMercury} improves agent performance across model scales and policy optimization methods.
Results are reported for both Qwen3.5-4B and Qwen3.5-35B-A3B models, with GRPO~\citep{shao2024deepseekmath} and SAO~\citep{hou2026single} considered as alternative RL algorithms. $\dagger$ signifies that the result are derived from \textsc{EnterpriseOps-Gym}. 
}
\resizebox{\textwidth}{!}{
\begin{tabular}{l ccccccccc}
\toprule
\textbf{Model}                & \textbf{Teams} & \textbf{CSM} & \textbf{Email} & \textbf{ITSM} & \textbf{Calendar} & \textbf{HR} & \textbf{Drive} & \textbf{Hybrid} & \textbf{Avg.} \\ \midrule
\textbf{GPT-5}~$\dagger$ & 26.3 & 36.4 & 49.0 & 18.9 & 41.3 & 17.9 & 34.0 & 23.5 & 30.9 \\
\textbf{Gemini-2.5-Pro}~$\dagger$ & 39.3 & 11.6 & 31.1 & 13.9 & 12.5 & 4.9 & 27.0 & 19.6 & 19.9 \\
\textbf{Kimi-K2-Thinking}~$\dagger$ &  30.0  & 7.1  & 51.0 &  12.2  & 15.4 & 8.2 & 39.6 & 15.7 & 22.4 \\
\textbf{Qwen3-4B (Think)}~$\dagger$ & 24.0 &  3.8 & 38.4 & 5.6 & 5.8 & 7.1 & 21.9 & 15.8 & 15.3 \\
\textbf{Qwen3-30B (Think)}~$\dagger$ & 22.0 & 5.4 & 51.9 & 6.7 & 18.3 & 7.6 & 25.7 & 15.7 & 19.1 \\
\textbf{Qwen3-235B (Think)}~$\dagger$ & 28.0 & 4.7 & 38.1 & 9.3 & 15.7 & 7.8 &  23.8 & 17.7 & 18.1\\
\midrule
\textbf{Qwen3.5-4B}                    & 20.8$\pm$2.5 & 9.2$\pm$2.3 & 23.9$\pm$3.0 & 6.8$\pm$1.0 & 10.4$\pm$2.5 & 9.5$\pm$2.8 &6.2$\pm$1.6& 11.7$\pm$1.7 &12.3$\pm$0.2\\ 
\textbf{Qwen3.5-4B + GRPO + \textsc{Ours}}      & 23.0$\pm$1.6 & 5.6$\pm$1.5 &33.3$\pm$0.9& 7.4$\pm$1.1 & 13.1$\pm$1.6 &  10.8$\pm$4.3  &15.6$\pm$3.1 & 17.0$\pm$1.1 & 15.7$\pm$0.6\\
\midrule
\textbf{Qwen3.5-35B-A3B}               &  33.9$\pm$3.8 & 7.6$\pm$2.5            & 53.2$\pm$2.3 & 14.6$\pm$1.0 & 18.6$\pm$0.9 & 11.4$\pm$1.5 & 35.9$\pm$5.4 & 23.5$\pm$1.7 & 24.8$\pm$0.6 \\
\textbf{Qwen3.5-35B-A3B + GRPO + \textsc{Ours}} & 39.8$\pm$4.1 & 11.1$\pm$0.9 & 53.8$\pm$7.5 & 15.6$\pm$1.6 & 21.9$\pm$1.9 & 17.7$\pm$0.9 & 41.2$\pm$3.9 & 23.7$\pm$1.1 & 28.1$\pm$1.4 \\
\textbf{Qwen3.5-35B-A3B + SAO + \textsc{Ours}}  & 39.1$\pm$2.3 & 12.5$\pm$2.8 & 54.8$\pm$2.1 & 16.5$\pm$0.5 & 22.6$\pm$0.5 & 17.7$\pm$1.4 & 39.1$\pm$4.4 & 24.1$\pm$2.4 & 28.3$\pm$1.5 \\
\bottomrule
\end{tabular}}{}
\label{tab:enterpriseops}
\vspace{-0.2cm}
\end{table}

\begin{table}[t]
\caption{Out-of-domain evaluation across mathematical reasoning, coding, scientific computing, tool use, and agentic benchmarks.
We report mean $\pm$ standard deviation over three independent evaluation runs ($N=3$).
Qwen3.5-4B and Qwen3.5-35B-A3B are evaluated before and after RL training on tasks synthesized by \textsc{AgentMercury}.
}
\resizebox{\textwidth}{!}{
\begin{tabular}{l ccccccccc}
\toprule
\textbf{Model}                & \textbf{AIME26} & \textbf{HMMT} & \textbf{LCB} & \textbf{SciCode} & \textbf{\begin{tabular}[c]{@{}c@{}}Tau3-\\ Airline\end{tabular}} & \textbf{\begin{tabular}[c]{@{}c@{}}Tau3-\\ Retail\end{tabular}} & \textbf{\begin{tabular}[c]{@{}c@{}}Tau3-\\ Telecom\end{tabular}} & \textbf{BFCL} & \textbf{\begin{tabular}[c]{@{}c@{}}GPQA-\\ Diamond\end{tabular}} \\ \midrule
\textbf{Qwen3.5-4B}                    & 45.9$\pm$1.7 & 28.5$\pm$1.2 & 36.6$\pm$2.8 & 22.6$\pm$0.1 & 48.8$\pm$2.1   & 70.4$\pm$1.0  &  92.5$\pm$1.0  & 30.3$\pm$1.8 & 76.5$\pm$0.7   \\ 
\textbf{Qwen3.5-4B + GRPO + \textsc{Ours}}      & 56.0$\pm$1.8 & 35.4$\pm$1.6 & 44.0$\pm$1.2 & 25.7$\pm$0.7 & 58.7$\pm$0.8 & 73.6$\pm$1.3 & 91.9$\pm$1.3 & 31.7$\pm$1.5 & 77.5$\pm$0.6\\ 
\midrule
\textbf{Qwen3.5-35B-A3B}               & 91.0$\pm$0.3 & 77.0$\pm$0.9 & 74.3$\pm$2.2 & 29.7$\pm$1.5 &  39.1$\pm$10.6   & 52.7$\pm$22.1   &  49.1$\pm$49.8   & 31.1$\pm$1.3 & 82.8 $\pm$3.0   \\ 
\textbf{Qwen3.5-35B-A3B + GRPO + \textsc{Ours}} &  91.9$\pm$2.1 &  80.0$\pm$0.7 & 79.0$\pm$0.5 & 29.9$\pm$1.7 &  52.5$\pm$5.2  & 57.1$\pm$8.3   & 68.9$\pm$20.5   & 42.5$\pm$0.3     &  85.0$\pm$0.5  \\ 
\textbf{Qwen3.5-35B-A3B + SAO + \textsc{Ours}}  & 92.2$\pm$1.9&       83.3$\pm$3.1      & 78.6$\pm$1.0 & 28.3$\pm$1.9 &  50.9$\pm$8.1 & 55.7$\pm$10.2   & 65.5$\pm$23.8   & 42.1$\pm$0.1 &  84.0$\pm$1.0  \\ \bottomrule
\end{tabular}}{}
\label{tab:overall_benchmark}
\end{table}

In Table~\ref{tab:enterpriseops}, our trained model, Qwen3.5-4B+GRPO+\textsc{Ours}, substantially improves over the base Qwen3.5-4B across most enterprise workflows.
With GRPO, the average score increases from $12.3$ to $15.7$, corresponding to a $+3.4$ point ($+27.6\%$) improvement.
The largest gains are observed in Drive, which improves from $6.2$ to $15.6$ ($+9.4$), and Email, which increases from $23.9$ to $33.3$ ($+9.4$).
We also observe notable improvements in Hybrid ($11.7\rightarrow17.0$, $+5.3$), Teams ($20.8\rightarrow23.0$, $+2.2$), and Calendar ($10.4\rightarrow13.1$, $+2.7$), while ITSM and HR improve by $+0.6$ and $+1.3$ points, respectively.
Although performance on CSM decreases from $9.2$ to $5.6$, the overall improvement across seven of the eight domains indicates that training on \textsc{AgentMercury}-synthesized environments transfers effectively to unseen enterprise workflows.

The improvement also extends beyond GRPO to SAO~\citep{hou2026single} at the larger model scale.
For Qwen3.5-35B-A3B, GRPO increases the average score from $24.8$ to $28.1$, a gain of $+3.3$ points ($+13.3\%$).
The improvement is consistent across all eight enterprise domains, with gains of $+5.9$ on Teams, $+3.5$ on CSM, $+2.5$ on Email, $+1.0$ on ITSM, $+3.0$ on Calendar, $+2.2$ on HR, $+5.3$ on Drive, and $+0.2$ on Hybrid.
Using SAO instead yields a comparable average score of $28.3$, corresponding to a $+3.5$ point ($+14.1\%$) improvement over the base model, again improving all eight domains.
In particular, SAO produces gains of $+5.2$ on Teams, $+4.9$ on CSM, $+1.6$ on Email, $+2.0$ on ITSM, $+3.9$ on Calendar, $+2.8$ on HR, $+3.2$ on Drive, and $+0.6$ on Hybrid.
These results suggest that the learning signal provided by \textsc{AgentMercury}-synthesized environments is not tied to a single policy optimization algorithm, but remains effective under both GRPO and SAO, particularly as the policy model scales.

\subsection{Policy Optimization of Benchmark-oriented Tasks}
As shown in Table~\ref{tab:overall_benchmark}, training on \textsc{AgentMercury} consistently improves the 4B and 35B model across substantially different capability domains.
In particular, Qwen3.5-4B + GRPO + \textsc{Ours} improves over the base model from $45.9$ to $56.0$ on AIME26, $28.5$ to $35.4$ on HMMT, and $36.6$ to $44.0$ on LiveCodeBench.
The same trend extends to scientific computing, with SciCode increasing from $22.6$ to $25.7$, and to structured tool use, where BFCL improves from $30.3$ to $31.7$.
We also observe gains on GPQA-Diamond ($76.5 \rightarrow 77.5$) and on two of the three $\tau^3$ domains: Airline improves from $48.8$ to $58.7$, while Retail improves from $70.4$ to $73.6$.
Telecom remains largely unchanged ($92.5 \rightarrow 91.9$), suggesting that the transfer is not simply a uniform reward-driven shift across all benchmarks.

The gains are particularly notable given that none of these benchmarks provides the model with the synthesized environments used during RL training.
The transfer from business-oriented tool-use trajectories to
mathematical reasoning, coding, scientific computing, and independent tool-use benchmarks therefore suggests that the synthesized tasks provide a broader policy-learning signal rather than merely inducing benchmark-specific behavior.

The effect is particularly pronounced for interactive tool-use
benchmarks.
While the Qwen3.5-35B-A3B base model achieves relatively strong mean performance on $\tau^3$, its results exhibit substantial run-to-run variance, especially on Telecom ($49.1\pm49.8$) and Retail ($52.7\pm22.1$).
After training with \textsc{AgentMercury} and SAO, the corresponding scores improve to $65.5\pm23.8$ and $55.7\pm10.2$, respectively, while Airline improves from $39.1\pm10.6$ to $50.9\pm8.1$.
Thus, the benefit of synthesized environments is not limited to raising the expected reward: the learned policy also becomes considerably more consistent across independently sampled interactions.
This stabilization is particularly important for agentic tasks, where successful behavior depends on a sequence of tool calls and state transitions rather than on producing a single correct answer.

At 35B scale, the gains are smaller on already-saturated reasoning benchmarks but remain substantial for tool-oriented capabilities.
For example, SAO improves AIME26 from $91.0$ to $92.2$, HMMT from $77.0$ to $83.3$, LiveCodeBench from $74.3$ to $78.6$, and BFCL from $31.1$ to $42.1$.
The particularly large BFCL gain suggests that the benefit of synthesized environments is especially pronounced when the target capability involves structured interaction with external tools.

\begin{figure}[t]
    \centering
    \includegraphics[width=\columnwidth]{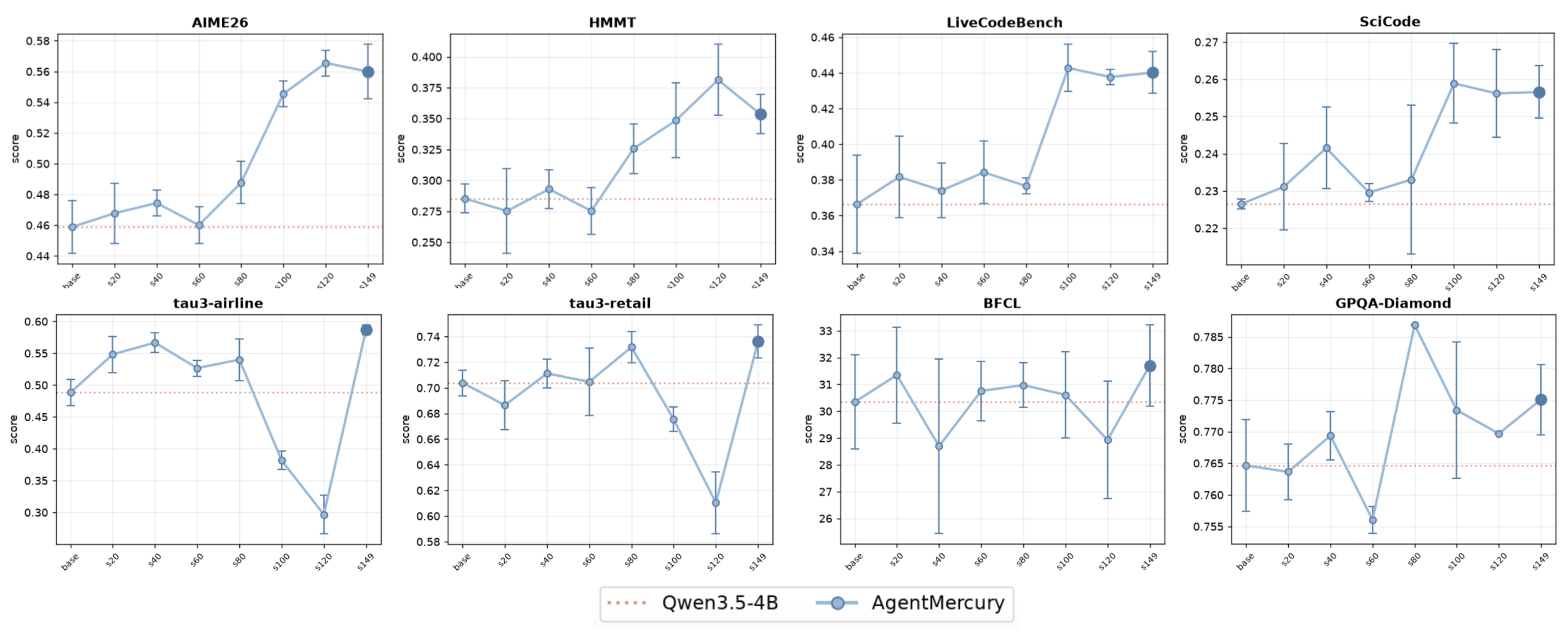}
    \caption{
    Out-of-domain benchmark performance across training checkpoints for Qwen3.5-4B + GRPO + \textsc{Ours}.
    The dashed line denotes the base-model performance, while error bars show the standard deviation over three independent evaluation runs.
    The gradual improvement across heterogeneous benchmarks indicates that the policy learned from \textsc{AgentMercury} environments acquires transferable capabilities beyond the training environments.
    }
    \label{fig:experiment_analysis}
    \vspace{-0.2cm}
\end{figure}

\begin{figure}[t]
    \centering
    \includegraphics[width=\columnwidth]{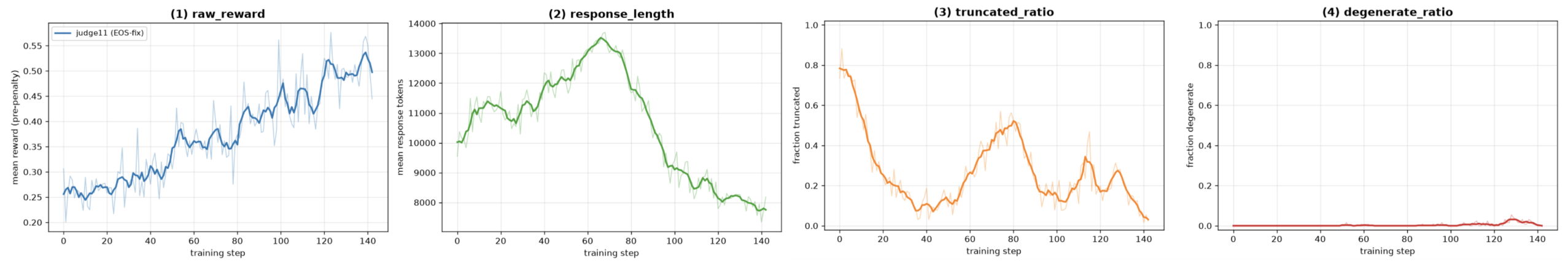}
    \caption{
    Training dynamics of Qwen3.5-4B + GRPO + \textsc{Ours}.
    We report the raw reward, response length, truncated-response ratio, and degenerate-response ratio over training steps.
    The increasing reward together with decreasing truncation and near-zero degeneration indicates that policy improvement is not accompanied by substantial response collapse or pathological generation.
    }
    \label{fig:training_log}
    \vspace{-0.2cm}
\end{figure}

Beyond the final checkpoint, we further examine how these improvements emerge during training.
Figure~\ref{fig:experiment_analysis} tracks the performance of
Qwen3.5-4B + GRPO + \textsc{Ours} across training checkpoints on the same set of out-of-domain benchmarks.
The improvement is progressive rather than being concentrated at a single checkpoint: mathematical reasoning, competitive programming, scientific computing, and tool-use capabilities generally improve as policy optimization proceeds.
In particular, AIME26 increases from the base score of $45.9$ to approximately $56.0$, while HMMT and LiveCodeBench similarly rise from $28.5$ to $35.4$ and from $36.6$ to $44.0$, respectively.
This checkpoint-level trend provides evidence that the final out-of-domain gains are a consequence of continued policy learning from the synthesized environments rather than evaluation noise at a single checkpoint.

We next examine the training dynamics to determine whether the observed capability gains are accompanied by pathological optimization behavior.
Figure~\ref{fig:training_log} shows the training trajectory of Qwen3.5-4B + GRPO + \textsc{Ours}.
The mean reward steadily increases throughout training, providing a direct indication that the policy increasingly satisfies the environment-level verification signals.
At the same time, the response length initially increases as the policy learns to solve more complex multi-step tasks, before gradually decreasing later in training.
Importantly, the truncated-response ratio decreases substantially over training, while the degenerate-response ratio remains near zero throughout.
Together, these trends indicate that the reward improvement is not driven by response degeneration or uncontrolled output growth, but by progressive optimization toward executable task completion.

Taken together, these results provide evidence that \textsc{AgentMercury} environments provide a useful and general learning signal for policy optimization.
Training on business-oriented synthesized environments improves performance not only on the corresponding enterprise workflows, but also on mathematical reasoning, coding, scientific computing, and independent tool-use benchmarks.
Moreover, the improvements emerge progressively during training and are accompanied by healthy optimization dynamics.
The reduction in variance on the larger model further suggests that the synthesized environments can improve the reliability of agent behavior, rather than merely increasing its average benchmark score.

\subsection{Can Agents Learn to Author Executable Worlds?}
\label{sec:exp_authoring}

Having established that synthesized worlds can serve as effective training
substrates, we next investigate whether the construction process itself can be
learned by an agent. Specifically, we ask whether a model given a high-level
business scenario can construct an executable world that satisfies the
required structural constraints, including services, state schemas, tools,
and cross-service invariants. This experiment treats environment authoring as
an agentic task with an explicit executable oracle: rather than judging the
generated world by surface-level similarity to a reference, we directly
execute the resulting environment and evaluate it against the structural
validators used by \textsc{AgentMercury}. This allows us to measure whether a
model has actually learned the construction process rather than merely
producing plausible-looking environment specifications.

We construct a held-out authoring set of 30 synthetic business briefs sampled
from the country--industry distribution of the environment library. Each brief
specifies a high-level business scenario without exposing the underlying
company environment used during construction. We evaluate five
off-the-shelf models in two settings: zero-shot, where the model is
given only the business brief, and recipe-conditioned, where the model
is additionally provided with an invariant digest and a trimmed example of a
previously constructed environment. The generated worlds are evaluated using
12 structural validators adapted from the \textsc{AgentMercury} construction
oracle. A world is counted as successful only when it passes all validators,
thereby requiring the generated environment to be executable and structurally
consistent rather than merely well-formed text.

\begin{table}[t]
\centering
\small
\caption{
Authoring executable worlds from high-level business briefs.
We report oracle-pass rates over 30 held-out briefs. A world is counted as successful only when it passes all 12 structural validators. Zero-shot provides only the business brief, while Recipe additionally provides an invariant digest and a trimmed construction exemplar.
}
\label{tab:authoring}
\begin{tabular}{@{}lcc@{}}
\toprule
Model & Zero-shot & Recipe \\
\midrule
Claude Opus 4.8 & 83.3\% & 80.0\% \\
GPT-5.4 & 66.7\% & 66.7\% \\
DeepSeek-V4-Pro & 83.3\% & 86.7\% \\
GLM-5.2 & 80.0\% & 76.7\% \\
\midrule
Mean (API models) & 78.3\% & 77.5\% \\
\midrule
Qwen3.5-35B-A3B & 3.3\% & 20.0\% \\
Qwen3.5-35B-A3B + \textsc{AgentMercury}
& \textbf{83.3\%} & 10.0\% \\
\bottomrule
\end{tabular}
\end{table}

\paragraph{Environment Authoring Performance.}
Table~\ref{tab:authoring} shows that executable-world authoring is feasible
for sufficiently capable models even without explicit construction examples.
The five API models achieve zero-shot oracle-pass rates between $66.7\%$ and
$90.0\%$, with a mean of $80.7\%$. Providing an explicit construction recipe
does not consistently improve performance: the mean pass rate decreases
slightly from $80.7\%$ to $78.0\%$, and the paired comparison shows no
systematic advantage for recipe conditioning. On the shared 30 briefs, the
two settings disagree on only a small fraction of examples for each model,
with the differences split in both directions. These results suggest that
the basic ability to construct structurally valid worlds is already present
in strong general-purpose models, and that simply exposing a construction
recipe is not sufficient to reliably improve this capability.

\paragraph{Failure Analysis.}
The remaining failures are concentrated in structural constraints that require
reasoning across multiple services rather than within an individual service.
In particular, a common failure mode is the collapse of a cross-service
constraint, where the trigger and target of an invariant are incorrectly
placed within the same service. This failure is especially pronounced for
GPT-5.4, for which 10 of the 30 briefs exhibit this error. Such failures are
important because they cannot be detected reliably from the generated
specification alone: the resulting world may appear syntactically valid while
violating the intended interaction structure between services. We therefore
view the executable oracle and its cross-service validators as an essential
part of the authoring process. Interestingly, recipe conditioning can even
increase formatting and parsing failures for some open models. 
This indicates that additional procedural information may
introduce a longer and more fragile generation format rather than directly
improving the underlying construction capability.

\paragraph{Learning the Authoring Process.}
The results change substantially when the construction traces themselves are
used as supervision. We fine-tune Qwen3.5-35B-A3B on 29,823 training samples
derived from the \textsc{AgentMercury} construction traces. The training set
contains complementary supervision for brief-to-world generation,
intermediate-stage completion, validator-guided corruption repair, and
intent-to-diff prediction. Before fine-tuning, the base model passes the full
oracle on only $3.3\%$ of the held-out briefs and frequently truncates or
breaks the required output format. After training, the oracle-pass rate
increases to $83.3\%$, matching the performance of the strongest
off-the-shelf API models. The improvement is accompanied by a substantial
increase in structural validity: the trained model passes an average of
$11.5$ out of the 12 validators, while truncation and formatting failures are
nearly eliminated. The difference is statistically significant, with a
Fisher exact test yielding $p=1.2\times10^{-10}$.

Interestingly, recipe conditioning behaves differently before and after
training. For the base Qwen3.5-35B-A3B model, providing the recipe improves
the oracle-pass rate from $3.3\%$ to $20.0\%$. In contrast, the same recipe
reduces the fine-tuned model's performance from $83.3\%$ to $10.0\%$. The
failure is highly concentrated: 27 of the 30 recipe-conditioned generations
from the fine-tuned model fail the cross-service validation check. This
contrast suggests that prompting and learning expose two qualitatively
different mechanisms. The recipe can partially compensate for the missing
construction capability of the base model, whereas after fine-tuning the
construction procedure is internalized in the model parameters and the
additional recipe can instead interfere with the learned generation policy.

These results provide evidence that environment authoring is
not merely an engineering procedure external to the agent. The construction
traces generated by \textsc{AgentMercury} contain sufficient supervision for
an agent to learn how to transform a high-level business scenario into an
executable and structurally valid world. This complements our first research
question: \textsc{AgentMercury} provides both \emph{worlds} that can be used
as training substrates and \emph{construction traces} that can teach agents
how to create such worlds. In this sense, the environment generator exposes
a learnable interface between high-level scenarios and executable agent
worlds, moving environment construction from a fixed engineering pipeline
toward an agent-learnable capability.

\section{Discussion and Conclusion}
\paragraph{What should we scale?}
A central observation from our experiments is that effective environment scaling does not necessarily require constructing environments around the target benchmarks.
\textsc{AgentMercury} generates environments from high-level business scenarios, without access to the task instances used by our evaluation benchmarks. Nevertheless, policies trained in these worlds improve not only on the business-oriented \textsc{EnterpriseOps-Gym}, but also on out-of-domain benchmarks spanning mathematical reasoning, coding, scientific computing, knowledge-intensive reasoning, and tool use.
This result suggests that the useful learning signal of an environment is not determined solely by its correspondence to a target benchmark.
Instead, diverse executable worlds can expose policies to reusable patterns of interaction, state tracking, constraint satisfaction, and tool-mediated decision making that transfer across task distributions.

This observation motivates a different perspective on environment scaling. 
Rather than scaling environments primarily by increasing the number of benchmark-specific tasks, we argue that future systems should investigate how to scale the diversity, structure, and realism of the underlying worlds.
In particular, business-oriented scenarios provide a natural source of long-horizon interactions, heterogeneous tools, persistent state, and cross-service constraints that are difficult to capture through benchmark-targeted task synthesis alone.

\paragraph{From benchmark optimization to world construction.}
Our results also suggest that the role of \textsc{Planet} may be more fundamental than that of a conventional data generator.
In the standard benchmark-centric paradigm, the environment is often treated as a means of producing more instances of a predefined task.
In contrast, \textsc{AgentMercury} separates the world, the task, and the policy: \textsc{Planet} determines what world exists, the task specifies what must be achieved within that world, and the policy learns how to act within it.
This factorization makes it possible to reuse a single synthesized world across multiple objectives and policy-training configurations, while also allowing the environment distribution itself to become a target of scaling.

We therefore view \textsc{Planet} as an important direction for building agents that are useful beyond benchmark optimization.
If environments can be scaled according to real-world scenarios rather than benchmark categories, the resulting policies may acquire capabilities that are useful for operating in open-ended settings even when those capabilities are not explicitly represented by a single evaluation suite.

\paragraph{Optimization dynamics across model scales.}
Our comparison between GRPO and SAO also reveals an interesting interaction between the policy optimization procedure and model scale.
As shown in Appendix~\ref{fig:35b_training_log} and Appendix~\ref{fig:35b_training_log_sao}, both methods can make progress on the synthesized environments at the 35B-A3B scale, but their optimization dynamics differ substantially. 
GRPO provides multiple sampled responses for the same prompt and can therefore obtain a relative learning signal even when the policy model is relatively small.
In contrast, we found that SAO was substantially less effective for the 4B model, where the available single-rollout signal was insufficiently represented in our experiments.
We therefore exclude the 4B SAO result from the main comparison rather than treating the unstable optimization as evidence against the underlying environment distribution.

At the same time, the behavior of SAO suggests a potentially useful scaling property.
Because SAO does not require multiple sampled responses from the same prompt, a single prompt can expose the policy to a broader set of environments and task instances. This property may become particularly attractive at larger model scales, where the objective is not only to optimize performance on a narrow task distribution but to inject diverse knowledge and interaction patterns from a large and heterogeneous
environment collection.
Thus, GRPO and SAO may occupy complementary points in the environment-learning trade-off: GRPO provides a stronger relative signal for smaller policies, whereas single-rollout optimization may offer a more scalable mechanism for broad environment coverage at larger scales.

\paragraph{Toward a closed environment-agent loop.}
Despite these results, our current system does not yet fully realize the closed loop between environment generation and agent learning.
In particular, the environment synthesis process does not currently employ a learned world model to predict how candidate worlds would behave under agent interaction. Consequently, the generator cannot yet use the policy's experience to identify which scenarios, tools, states, or interaction patterns are most needed for further learning and then synthesize those worlds accordingly.

A natural next step is therefore to connect the \textsc{Planet} role more tightly with policy learning.
An agent could monitor its own failures and uncertainties, identify underrepresented capabilities or interaction patterns, and propose new high-level scenarios targeted at those gaps.
These scenarios could then be compiled into executable worlds, evaluated by deterministic verifiers, and fed back into policy optimization.
Such a loop would turn environment synthesis from a largely offline data generation process into an adaptive curriculum in which the policy actively determines what worlds should be generated next.
A learned world model such as Qwen-AgentWorld could further support this process by predicting the consequences of candidate environments before they are fully instantiated, enabling more efficient search over the environment space.

\paragraph{Conclusion.}
\textsc{AgentMercury} explores a shift in how training environments for agents are constructed: rather than scaling isolated task instances around fixed benchmarks, we scale persistent, executable worlds grounded in high-level business scenarios.
We synthesize 4,783 environments spanning 14 industries and 50 countries, each with persistent state, multi-service interactions, and executable cross-service constraints.
These worlds provide effective training signals for policy learning: Qwen3.5-4B trained with GRPO improves from $12.3$ to $15.7$ on EnterpriseOps-GYM, while also improving substantially on out-of-domain benchmarks such as AIME26 ($45.9\rightarrow56.0$), HMMT ($28.5\rightarrow35.4$), LiveCodeBench ($36.6\rightarrow44.0$), and SciCode ($22.6\rightarrow25.7$), despite the training environments being constructed independently of these evaluation tasks.
We further show that the construction process itself can be learned: fine-tuning Qwen3.5-35B-A3B on construction traces increases executable-world authoring success from $3.3\%$ to $83.3\%$ on held-out business scenarios.
Together, these results show that scenario-grounded environments can serve both as scalable training substrates for policies and as structured supervision for learning how executable worlds are constructed.

More broadly, our results suggest that environment scaling need not be defined by generating increasingly many instances of increasingly difficult benchmarks.
Instead, scaling the diversity and structure of the underlying worlds can provide learning signals that transfer across tasks and capabilities.
This motivates treating \textsc{Planet} as a first-class, learnable component of the agent system: while policies learn how to act within a world and world models learn how the world responds, \textsc{Planet} can determine which worlds should exist and how they should evolve.
Our current system does not yet close this loop, environment synthesis is not driven by a learned world model that identifies a policy's capability gaps and generates the next most useful scenarios.
We view this as an important direction for future work, toward a continual process in which agents act in executable worlds, identify what they still need to learn, and \textsc{Planet} constructs new worlds to provide that experience.
In this view, the goal is not merely to build better environments for today's benchmarks, but to develop increasingly diverse and useful worlds in which agents can learn capabilities that matter beyond the benchmarks themselves.

\bibliography{iclr2026_conference}

@inproceedings{taubench,
  title         = {{$\tau$-bench}: A Benchmark for Tool-Agent-User Interaction in Real-World Domains},
  author        = {Yao, Shunyu and Shinn, Noah and Razavi, Pedram and Narasimhan, Karthik},
  booktitle     = {The Thirteenth International Conference on Learning Representations (ICLR)},
  year          = {2025},
  note          = {arXiv:2406.12045}
}

@misc{tau2bench,
  title         = {{$\tau^2$-Bench}: Evaluating Conversational Agents in a Dual-Control Environment},
  author        = {Barres, Victor and Dong, Honghua and Ray, Soham and Si, Xujie and Narasimhan, Karthik},
  year          = {2025},
  note          = {arXiv:2506.07982},
  url           = {https://arxiv.org/abs/2506.07982}
}

@article{bandi2026mcp,
  title={Mcp-atlas: A large-scale benchmark for tool-use competency with real mcp servers},
  author={Bandi, Chaithanya and Dumitru, Razvan-Gabriel and Hertzberg, Ben and Agarwal, Divyansh and Boo, Geobio and Polakam, Tejas and Hassaan, Sami and Da, Jeff and Kim, HiJae and Gupta, Vipul and others},
  journal={arXiv preprint arXiv:2602.00933},
  year={2026}
}

@inproceedings{appworld,
  title         = {{AppWorld}: A Controllable World of Apps and People for Benchmarking Interactive Coding Agents},
  author        = {Trivedi, Harsh and Khot, Tushar and Hartmann, Mareike and Manku, Ruskin and Dong, Vinty and Li, Edward and Gupta, Shashank and Sabharwal, Ashish and Balasubramanian, Niranjan},
  booktitle     = {Proceedings of the 62nd Annual Meeting of the Association for Computational Linguistics (Volume 1: Long Papers)},
  year          = {2024},
  pages         = {16022--16076},
  note          = {Best Resource Paper; arXiv:2407.18901}
}

@article{zeng2026glm,
  title={Glm-5: from vibe coding to agentic engineering},
  author={Zeng, Aohan and Lv, Xin and Hou, Zhenyu and Du, Zhengxiao and Zheng, Qinkai and Chen, Bin and Yin, Da and Ge, Chendi and Huang, Chenghua and Xie, Chengxing and others},
  journal={arXiv preprint arXiv:2602.15763},
  year={2026}
}

@inproceedings{wu2026mcpmark,
  title={Mcpmark: A benchmark for stress-testing realistic and comprehensive mcp use},
  author={Wu, Zijian and Liu, Xiangyan and Chen, Lingjun and Meng, Fanqing and Du, Lingxiao and Zhao, Yiran and Zhang, Fanshi and Ye, Yaoqi and Wang, Jiawei and Wang, Zirui and others},
  booktitle={International Conference on Learning Representations},
  year={2026}
}

@article{zheng2026bench,
  title={E-Bench: Benchmarking Multi-Step Tool-Use Agents in Real-World Product Scenarios},
  author={Zheng, Weihuang and Zou, Tianyuan and Ye, Eileen and Liu, Alphet and Kong, Youyong and Zhang, Ya-Qin and Zheng, Duran and Pan, Maxm},
  journal={arXiv preprint arXiv:2607.23722},
  year={2026}
}

@inproceedings{toolathlon,
  title         = {{The Tool Decathlon}: Benchmarking Language Agents for Diverse, Realistic, and Long-Horizon Task Execution},
  author        = {Li, Junlong and Zhao, Wenshuo and Zhao, Jian and Zeng, Weihao and Wu, Haoze and Wang, Xiaochen and Ge, Rui and Cao, Yuxuan and Huang, Yuzhen and Liu, Wei and Liu, Junteng and Su, Zhaochen and Guo, Yiyang and Zhou, Fan and Zhang, Lueyang and Michelini, Juan and Wang, Xingyao and Yue, Xiang and Zhou, Shuyan and Neubig, Graham and He, Junxian},
  booktitle     = {The Fourteenth International Conference on Learning Representations (ICLR)},
  year          = {2026},
  eprint        = {2510.25726},
  archivePrefix = {arXiv}
}

@article{mavali2026no,
  title={No More, No Less: Task Alignment in Terminal Agents},
  author={Mavali, Sina and Pape, David and Evertz, Jonathan and Abedini, Samira and Srivastav, Devansh and Eisenhofer, Thorsten and Abdelnabi, Sahar and Sch{\"o}nherr, Lea},
  journal={arXiv preprint arXiv:2605.12233},
  year={2026}
}

@inproceedings{jimenez2024swe,
  title={Swe-bench: Can language models resolve real-world github issues?},
  author={Jimenez, Carlos E and Yang, John and Wettig, Alexander and Yao, Shunyu and Pei, Kexin and Press, Ofir and Narasimhan, Karthik},
  booktitle={International Conference on Learning Representations},
  year={2024}
}

@article{team2026kimi,
  title={Kimi K3: Open Frontier Intelligence},
  author={Team, Kimi and Bai, Tongtong and Bai, Yifan and Bao, Yiping and Cai, Jianfeng and Cai, Xinyuan and Cao, Peizhou and Cao, Yuxuan and Chai, Ziwei and Charles, Y and others},
  journal={arXiv preprint arXiv:2607.24653},
  year={2026}
}

@inproceedings{bfcl,
  title         = {The {Berkeley} Function Calling Leaderboard ({BFCL}): From Tool Use to Agentic Evaluation of Large Language Models},
  author        = {Patil, Shishir G. and Mao, Huanzhi and Yan, Fanjia and Ji, Charlie Cheng-Jie and Suresh, Vishnu and Stoica, Ion and Gonzalez, Joseph E.},
  booktitle     = {Proceedings of the 42nd International Conference on Machine Learning (ICML)},
  year          = {2025},
  note          = {PMLR 267:48371--48392}
}

@article{dong2026agent,
  title={Agent-world: Scaling real-world environment synthesis for evolving general agent intelligence},
  author={Dong, Guanting and Lu, Junting and Huang, Junjie and Zhong, Wanjun and Liu, Longxiang and Huang, Shijue and Li, Zhenyu and Zhao, Yang and Song, Xiaoshuai and Li, Xiaoxi and others},
  journal={arXiv preprint arXiv:2604.18292},
  year={2026}
}

@article{ha2018world,
  title={World models},
  author={Ha, David and Schmidhuber, J{\"u}rgen},
  journal={arXiv preprint arXiv:1803.10122},
  year={2018}
}

@article{zuo2026qwen,
  title={Qwen-AgentWorld: Language World Models for General Agents},
  author={Zuo, Yuxin and Xiao, Zikai and Sheng, Li and Huang, Fei and Tu, Jianhong and Liu, Yuxuan and Tang, Tianyi and Hu, Xiaomeng and Su, Yang and Lan, Qingfeng and others},
  journal={arXiv preprint arXiv:2606.24597},
  year={2026}
}

@inproceedings{toolllm,
  title         = {{ToolLLM}: Facilitating Large Language Models to Master 16000+ Real-world {APIs}},
  author        = {Qin, Yujia and Liang, Shihao and Ye, Yining and Zhu, Kunlun and Yan, Lan and Lu, Yaxi and Lin, Yankai and Cong, Xin and Tang, Xiangru and Qian, Bill and Zhao, Sihan and Hong, Lauren and Tian, Runchu and Xie, Ruobing and Zhou, Jie and Gerstein, Mark and Li, Dahai and Liu, Zhiyuan and Sun, Maosong},
  booktitle     = {The Twelfth International Conference on Learning Representations (ICLR)},
  year          = {2024},
  note          = {arXiv:2307.16789}
}

@misc{terminalworld,
  title         = {{Terminal-World}: Scaling Terminal-Agent Environments via Agent Skills},
  author        = {Cheng, Zihao and Wang, Hongru and Liu, Zeming and Wang, Xinyi and Zhu, Xiangrong and Guo, Yuhang and Lin, Wei and Pan, Jeff Z. and Wang, Yunhong},
  year          = {2026},
  note          = {arXiv:2605.20876}
}

@article{phoneworld,
  title         = {{PhoneWorld}: Scaling Phone-Use Agent Environments},
  author        = {{Tencent Hunyuan}},
  journal       = {arXiv preprint arXiv:2605.29486},
  year          = {2026}
}

@inproceedings{react,
  title={{ReAct}: Synergizing Reasoning and Acting in Language Models},
  author={Yao, Shunyu and Zhao, Jeffrey and Yu, Dian and Du, Nan and Shafran, Izhak and Narasimhan, Karthik and Cao, Yuan},
  year={2023},
  booktitle={International Conference on Learning Representations (ICLR)}
}

@article{nakano2021webgpt,
  title={Webgpt: Browser-assisted question-answering with human feedback},
  author={Nakano, Reiichiro and Hilton, Jacob and Balaji, Suchir and Wu, Jeff and Ouyang, Long and Kim, Christina and Hesse, Christopher and Jain, Shantanu and Kosaraju, Vineet and Saunders, William and others},
  journal={arXiv preprint arXiv:2112.09332},
  year={2021}
}

@inproceedings{cobbe2020leveraging,
  title={Leveraging procedural generation to benchmark reinforcement learning},
  author={Cobbe, Karl and Hesse, Chris and Hilton, Jacob and Schulman, John},
  booktitle={International conference on machine learning},
  pages={2048--2056},
  year={2020},
  organization={PMLR}
}

@article{schrittwieser2020mastering,
  title={Mastering atari, go, chess and shogi by planning with a learned model},
  author={Schrittwieser, Julian and Antonoglou, Ioannis and Hubert, Thomas and Simonyan, Karen and Sifre, Laurent and Schmitt, Simon and Guez, Arthur and Lockhart, Edward and Hassabis, Demis and Graepel, Thore and others},
  journal={Nature},
  volume={588},
  number={7839},
  pages={604--609},
  year={2020},
  publisher={Nature Publishing Group UK London}
}

@article{hafner2023mastering,
  title={Mastering diverse domains through world models},
  author={Hafner, Danijar and Pasukonis, Jurgis and Ba, Jimmy and Lillicrap, Timothy},
  journal={arXiv preprint arXiv:2301.04104},
  year={2023}
}

@article{hafner2019dream,
  title={Dream to control: Learning behaviors by latent imagination},
  author={Hafner, Danijar and Lillicrap, Timothy and Ba, Jimmy and Norouzi, Mohammad},
  journal={arXiv preprint arXiv:1912.01603},
  year={2019}
}

@inproceedings{apibank,
  title={{API-Bank}: A Comprehensive Benchmark for Tool-Augmented {LLMs}},
  author={Li, Minghao and Zhao, Yingxiu and Yu, Bowen and Song, Feifan and Li, Hangyu and Yu, Haiyang and Li, Zhoujun and Huang, Fei and Li, Yongbin},
  year={2023},
  booktitle={Proceedings of the 2023 Conference on Empirical Methods in Natural Language Processing (EMNLP)}
}

@inproceedings{webvoyager,
  title={{WebVoyager}: Building an End-to-End Web Agent with Large Multimodal Models},
  author={He, Hongliang and Yao, Wenlin and Ma, Kaixin and Yu, Wenhao and Dai, Yong and Zhang, Hongming and Lan, Zhenzhong and Yu, Dong},
  year={2024},
  booktitle={Proceedings of the 62nd Annual Meeting of the Association for Computational Linguistics (ACL)}
}

@inproceedings{alfworld,
  title={{ALFWorld}: Aligning Text and Embodied Environments for Interactive Learning},
  author={Shridhar, Mohit and Yuan, Xingdi and C{\^o}t{\'e}, Marc-Alexandre and Bisk, Yonatan and Trischler, Adam and Hausknecht, Matthew},
  year={2021},
  booktitle={International Conference on Learning Representations (ICLR)}
}

@book{sutton1998reinforcement,
  title={Reinforcement learning: An introduction},
  author={Sutton, Richard S and Barto, Andrew G and Barto, Andrew},
  year={1998},
  publisher={MIT press Cambridge}
}

@article{hou2026single,
  title={Single-Rollout Asynchronous Optimization for Agentic Reinforcement Learning},
  author={Hou, Zhenyu and Li, Yujiang and Tang, Jie and Dong, Yuxiao},
  journal={arXiv preprint arXiv:2607.07508},
  year={2026}
}

@misc{deepseekai2026deepseekv4,
      title={DeepSeek-V4: Towards Highly Efficient Million-Token Context Intelligence},
      author={DeepSeek-AI},
      year={2026},
}

@article{malay2026enterpriseops,
  title={Enterpriseops-gym: Environments and evaluations for stateful agentic planning and tool use in enterprise settings},
  author={Malay, Shiva Krishna Reddy and Nayak, Shravan and Nair, Jishnu Sethumadhavan and Davasam, Sagar and Tiwari, Aman and Madhusudhan, Sathwik Tejaswi and Nemala, Sridhar Krishna and Sunkara, Srinivas and Rajeswar, Sai},
  journal={arXiv preprint arXiv:2603.13594},
  year={2026}
}

@article{deng2025swe,
  title={Swe-bench pro: Can ai agents solve long-horizon software engineering tasks?},
  author={Deng, Xiang and Da, Jeff and Pan, Edwin and He, Yannis Yiming and Ide, Charles and Garg, Kanak and Lauffer, Niklas and Park, Andrew and Pasari, Nitin and Rane, Chetan and others},
  journal={arXiv preprint arXiv:2509.16941},
  year={2025}
}

@article{xu2026theagentcompany,
  title={Theagentcompany: benchmarking llm agents on consequential real world tasks},
  author={Xu, Frank Fangzheng and Song, Yufan and Li, Boxuan and Tang, Yuxuan and Jain, Kritanjali and Bao, Mengxue and Wang, Zora and Zhou, Xuhui and Guo, Zhitong and Cao, Murong and others},
  journal={Advances in Neural Information Processing Systems},
  year={2026}
}

@article{xu2026envfactory,
  title={EnvFactory: Scaling Tool-Use Agents via Executable Environments Synthesis and Robust RL},
  author={Xu, Minrui and Wang, Zilin and Deng, Mengyi and Li, Zhiwei and Yang, Zhicheng and Zhu, Xiao and Liu, Yinhong and Zhu, Boyu and Huang, Baiyu and Chen, Chao and others},
  journal={arXiv preprint arXiv:2605.18703},
  year={2026}
}

@inproceedings{song2026envscaler,
  title={Envscaler: Scaling tool-interactive environments for llm agent via programmatic synthesis},
  author={Song, Xiaoshuai and Chang, Haofei and Dong, Guanting and Zhu, Yutao and Wen, Ji-Rong and Dou, Zhicheng},
  booktitle={Findings of the Association for Computational Linguistics: ACL 2026},
  year={2026}
}

@article{wang2026agent,
  title={Agent world model: Infinity synthetic environments for agentic reinforcement learning},
  author={Wang, Zhaoyang and Xu, Canwen and Liu, Boyi and Wang, Yite and Han, Siwei and Yao, Zhewei and Yao, Huaxiu and He, Yuxiong},
  journal={arXiv preprint arXiv:2602.10090},
  year={2026}
}

@inproceedings{liu2024agentbench,
  title={Agentbench: Evaluating llms as agents},
  author={Liu, Xiao and Yu, Hao and Zhang, Hanchen and Xu, Yifan and Lei, Xuanyu and Lai, Hanyu and Gu, Yu and Ding, Hangliang and Men, Kaiwen and Yang, Kejuan and others},
  booktitle={International Conference on Learning Representations},
  year={2024}
}

@article{yao2022webshop,
  title={Webshop: Towards scalable real-world web interaction with grounded language agents},
  author={Yao, Shunyu and Chen, Howard and Yang, John and Narasimhan, Karthik},
  journal={Advances in Neural Information Processing Systems},
  year={2022}
}

@article{drouin2024workarena,
  title={WorkArena: How capable are web agents at solving common knowledge work tasks?},
  author={Drouin, Alexandre and Gasse, Maxime and Caccia, Massimo and Laradji, Issam H and Del Verme, Manuel and Marty, Tom and Boisvert, L{\'e}o and Thakkar, Megh and Cappart, Quentin and Vazquez, David and others},
  journal={arXiv preprint arXiv:2403.07718},
  year={2024}
}

@article{xie2024osworld,
  title={Osworld: Benchmarking multimodal agents for open-ended tasks in real computer environments},
  author={Xie, Tianbao and Zhang, Danyang and Chen, Jixuan and Li, Xiaochuan and Zhao, Siheng and Cao, Ruisheng and Hua, Toh J and Cheng, Zhoujun and Shin, Dongchan and Lei, Fangyu and others},
  journal={Advances in Neural Information Processing Systems},
  year={2024}
}

@inproceedings{zhou2024webarena,
  title={Webarena: A realistic web environment for building autonomous agents},
  author={Zhou, Shuyan and Xu, Frank F and Zhu, Hao and Zhou, Xuhui and Lo, Robert and Sridhar, Abishek and Cheng, Xianyi and Ou, Tianyue and Bisk, Yonatan and Fried, Daniel and others},
  booktitle={International Conference on Learning Representations},
  year={2024}
}

@misc{aime26,
  title={American Invitational Mathematics Examination},
  author={Mathematical Association of America},
  year={2026},
  note={https://maa.org/maa-invitational-competitions/}
}

@article{tian2024scicode,
  title={Scicode: A research coding benchmark curated by scientists},
  author={Tian, Minyang and Gao, Luyu and Zhang, Shizhuo D and Chen, Xinan and Fan, Cunwei and Guo, Xuefei and Haas, Roland and Ji, Pan and Krongchon, Kittithat and Li, Yao and others},
  journal={Advances in Neural Information Processing Systems},
  year={2024}
}

@article{rein2023gpqa,
  title={Gpqa: A graduate-level google-proof q\&a benchmark},
  author={Rein, David and Hou, Betty Li and Stickland, Asa Cooper and Petty, Jackson and Pang, Richard Yuanzhe and Dirani, Julien and Michael, Julian and Bowman, Samuel R},
  journal={arXiv preprint arXiv:2311.12022},
  year={2023}
}

@inproceedings{patil2025berkeley,
  title={The berkeley function calling leaderboard (bfcl): From tool use to agentic evaluation of large language models},
  author={Patil, Shishir G and Mao, Huanzhi and Yan, Fanjia and Ji, Charlie Cheng-Jie and Suresh, Vishnu and Stoica, Ion and Gonzalez, Joseph E},
  booktitle={Forty-second International Conference on Machine Learning},
  year={2025}
}

@article{shi2026tau,
  title={tau-Knowledge: Evaluating Conversational Agents over Unstructured Knowledge},
  author={Shi, Quan and Zytek, Alexandra and Razavi, Pedram and Narasimhan, Karthik and Barres, Victor},
  journal={arXiv preprint arXiv:2603.04370},
  year={2026}
}

@inproceedings{jain2025livecodebench,
  title={Livecodebench: Holistic and contamination free evaluation of large language models for code},
  author={Jain, Naman and Gu, Alex and Li, Wen-Ding and Yan, Fanjia and Zhang, Tianjun and Wang, Sida and Solar-Lezama, Armando and Sen, Koushik and Stoica, Ion},
  booktitle={International Conference on Learning Representations},
  volume={2025},
  year={2025}
}

@article{dekoninck2026matharena,
      title={Beyond Benchmarks: MathArena as an Evaluation Platform for Mathematics with LLMs}, 
      author={Jasper Dekoninck and Nikola Jovanović and Tim Gehrunger and Kári Rögnvaldsson and Ivo Petrov and Chenhao Sun and Martin Vechev},
      year={2026},
      eprint={2605.00674},
      url={https://arxiv.org/abs/2605.00674}, 
}

@misc{qwen3.5,
    title  = {{Qwen3.5}: Towards Native Multimodal Agents},
    author = {{Qwen Team}},
    month  = {February},
    year   = {2026},
    url    = {https://qwen.ai/blog?id=qwen3.5}
}

@article{shao2024deepseekmath,
  title={Deepseekmath: Pushing the limits of mathematical reasoning in open language models},
  author={Shao, Zhihong and Wang, Peiyi and Zhu, Qihao and Xu, Runxin and Song, Junxiao and Bi, Xiao and Zhang, Haowei and Zhang, Mingchuan and Li, YK and Wu, Yang and others},
  journal={arXiv preprint arXiv:2402.03300},
  year={2024}
}

@article{liu2025understanding,
  title={Understanding r1-zero-like training: A critical perspective},
  author={Liu, Zichen and Chen, Changyu and Li, Wenjun and Qi, Penghui and Pang, Tianyu and Du, Chao and Lee, Wee Sun and Lin, Min},
  journal={arXiv preprint arXiv:2503.20783},
  year={2025}
}
\bibliographystyle{iclr2026_conference}

\appendix

\section{Synthesized \& Sampled Tasks}
\label{app:tasks}

\paragraph{Task Instantiation from Synthesized Worlds.}
\textsc{AgentMercury} separates world construction from task
instantiation. Once an executable world $w$ has been synthesized,
multiple tasks can be instantiated from the same underlying world by
varying the seeded state, user intent, and task rubric. In our released
RL corpus, each task corresponds to an autonomous investigation
instance associated with a synthetic company environment and a
grounded persona. Thus, the task distribution is sampled on top of a
shared distribution of executable worlds rather than being used as the
primary specification for constructing those worlds.

Formally, given a synthesized world
$w = \langle \mathcal{S},\mathcal{A},\Omega,T,O,s_0,\mathcal{R}\rangle$,
we instantiate a task as
\begin{equation}
    (\Delta s_0,u,\rho)
    \sim
    \operatorname{Task}(\cdot \mid s_0,\mathcal{R}),
\end{equation}
where $\Delta s_0$ specifies task-specific state seeding, $u$ is the natural-language user instruction, and $\rho$ is the task-level rubric.
The resulting task therefore modifies the initial state and specifies the objective and evaluation criteria without changing the underlying world structure.
This separation allows the same world to support multiple task instances and interaction trajectories.

\begin{figure*}[t]
    \centering
    \includegraphics[width=\columnwidth]{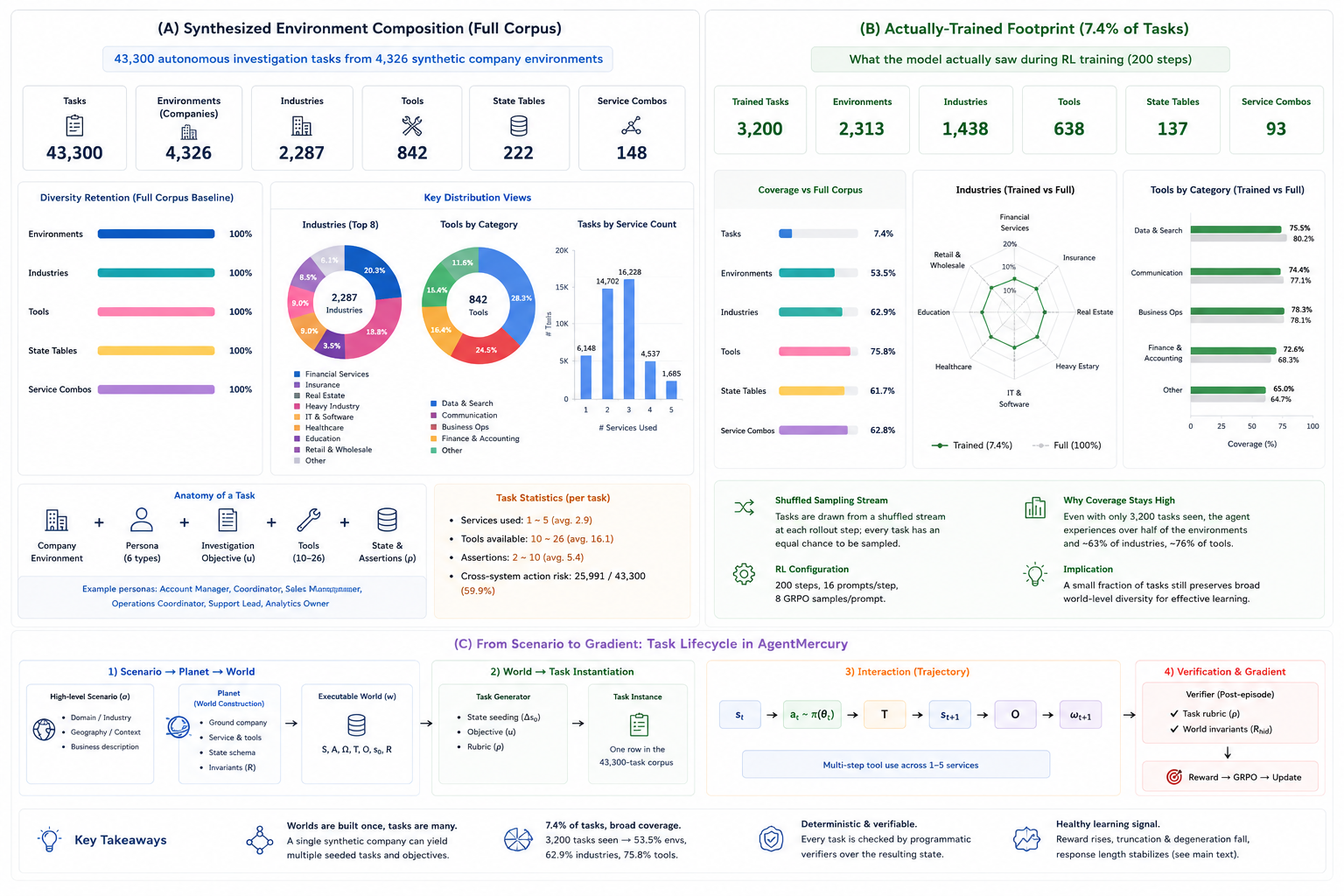}
    \caption{
    Overview of the synthesized environments, sampled training footprint, and task lifecycle in \textsc{AgentMercury}. (A) The full corpus contains 43,300 investigation tasks instantiated from 4,326 synthetic company environments, spanning 2,287 industries, 842 tools, 222 state tables, and 148 service combinations.
    (B) Across 200 RL training steps, the agent observes 3,200 tasks (7.4\% of the full task corpus) while retaining broad coverage of environments, industries, tools, state tables, and service combinations through shuffled sampling.
    (C) Each training instance follows a structured lifecycle from a high-level business scenario to executable world construction, task instantiation, multi-step tool interaction, deterministic post-episode verification, and reward-based policy optimization.
    The resulting framework separates world construction from task sampling, allowing a single synthesized environment to support multiple seeded tasks while providing verifiable learning signals for agent training.
    }
    \label{fig:task_synthesis}
    \vspace{-0.2cm}
\end{figure*}

\paragraph{Corpus Composition.}
In Figure~\ref{fig:task_synthesis}, we depict overall statistics of our corpus composition. 
The RL training corpus used in our experiments contains
$43{,}300$ synthesized tasks collected from $4{,}326$ executable company
environments, with ten sampled task seeds per environment.
The environments span $2{,}287$ distinct industry descriptions and
expose $842$ unique tools over $222$ state-table types and
$148$ distinct service combinations. Each environment exposes between
10 and 26 tools, with a mean of 16.1 tools, while each task touches
between one and five services, with a mean of 2.9 services.
Each task contains a mean of 5.4 programmatic assertions, ranging from
2 to 10 assertions. These statistics are computed over the complete
43,300-task corpus. 

\begin{table}[t]
\centering
\small
\caption{Composition of the synthesized task corpus used for RL.
Statistics are computed over all 43,300 tasks.}
\label{tab:task_corpus}
\begin{tabular}{@{}lr@{}}
\toprule
Statistic & Value \\
\midrule
Tasks & 43,300 \\
Executable worlds / companies & 4,326 \\
Task seeds per world & 10 \\
Unique industries & 2,287 \\
Unique tools & 842 \\
Unique state tables & 222 \\
Service combinations & 148 \\
Tools per world & 10--26 (16.1 avg.) \\
Services per task & 1--5 (2.9 avg.) \\
Assertions per task & 2--10 (5.4 avg.) \\
\bottomrule
\end{tabular}
\end{table}

\paragraph{Task Structure.}
Each task combines a synthetic company environment with a user persona,
a natural-language investigation objective, executable tools, and
programmatic state-based verification. The task instances in the RL
corpus use six primary persona roles: Account Manager, Coordinator,
Sales Manager, Operations Coordinator, Support Lead, and Analytics
Owner. The most common service combinations involve CRM, email, and
Slack, while additional tasks introduce billing, ticketing, support,
project-management, and domain-specific systems. Overall, the corpus
contains 148 distinct service combinations, providing variation in both
the number and identity of systems involved in a task.

The task objectives are grounded in the state of the corresponding
environment. An agent must therefore discover relevant information
through tool interaction and, when required, modify the underlying
state. Rather than evaluating only the textual response, task
completion is checked against the resulting database state using
programmatic assertions. Across the corpus, the assertions comprise
three basic forms: record existence, field equality, and field
inequality. This provides a deterministic signal for whether the
intended state change has actually occurred.

\paragraph{Task Difficulty and Cross-Service Interaction.}
The sampled tasks vary substantially in interaction breadth and
difficulty. The number of services touched by a task ranges from one
to five, with three-service tasks constituting the largest group.
In addition, tasks are annotated with up to six hard requirements that
capture more demanding forms of business reasoning, such as precise
scoping, reconciliation, and cross-system correction. Across the full
corpus, 25,991 of 43,300 tasks are flagged as involving
cross-system action risk, indicating that the required actions may
propagate across multiple business systems.

This structure is important for RL because task diversity is not
obtained solely by changing the textual instruction. Different tasks
can expose the policy to different combinations of services, tools,
state tables, personas, and cross-system dependencies, while retaining
the same executable semantics of the underlying world.

\paragraph{Sampling During RL.}
Although the complete corpus contains 43,300 tasks, each RL run
observes only a subset of these tasks through the rollout sampler.
For the reported Qwen3.5-4B RL run, 3,200 unique task instances
received gradients over 200 training steps, corresponding to
approximately $7.4\%$ of the full corpus. The sampled tasks were drawn
from a shuffled task stream rather than from a manually selected
subset.

Importantly, the relatively small number of tasks used directly for
optimization does not imply a narrow training distribution. The
3,200-task subset spans 2,313 distinct environments, 1,438 industries,
638 unique tools, 137 state tables, and 93 service combinations.
Thus, although only $7.4\%$ of task instances received gradients, the
training subset retains a substantial fraction of the structural
diversity present in the full corpus.

    \begin{table}[t]
\centering
\small
\caption{Coverage of the tasks that received gradients during the reported RL run relative to the full synthesized corpus. These statistics used only for GRPO algorithm.}
\label{tab:task_sampling}
\begin{tabular}{@{}lrrr@{}}
\toprule
Statistic & Full & Trained & Coverage \\
\midrule
Tasks & 43,300 & 3,200 & 7.4\% \\
Environments & 4,326 & 2,313 & 53.5\% \\
Industries & 2,287 & 1,438 & 62.9\% \\
Tools & 842 & 638 & 75.8\% \\
State tables & 222 & 137 & 61.7\% \\
Service combinations & 148 & 93 & 62.8\% \\
\bottomrule
\end{tabular}
\end{table}

\paragraph{Implication for Training Diversity.}
These statistics illustrate the distinction between \emph{task
count} and \emph{world diversity}. A small fraction of the task corpus
can still cover a large fraction of the underlying environments,
industries, tools, and state structures when tasks are sampled across
the synthesized world distribution. In our RL experiment, the
3,200-task training footprint therefore represents a broad
cross-section of the synthesized worlds rather than a narrow set of
repeated tasks.

This sampling procedure is a direct consequence of the
scenario-grounded construction paradigm: \textsc{AgentMercury} first
creates executable worlds and subsequently samples tasks from those
worlds. The resulting training distribution can consequently expand along multiple axes---new worlds, new business structures, new service combinations, new initial states, and new task objectives--- without requiring a separate manually constructed environment for each task.

\begin{figure}[t]
    \centering
    \includegraphics[width=\columnwidth]{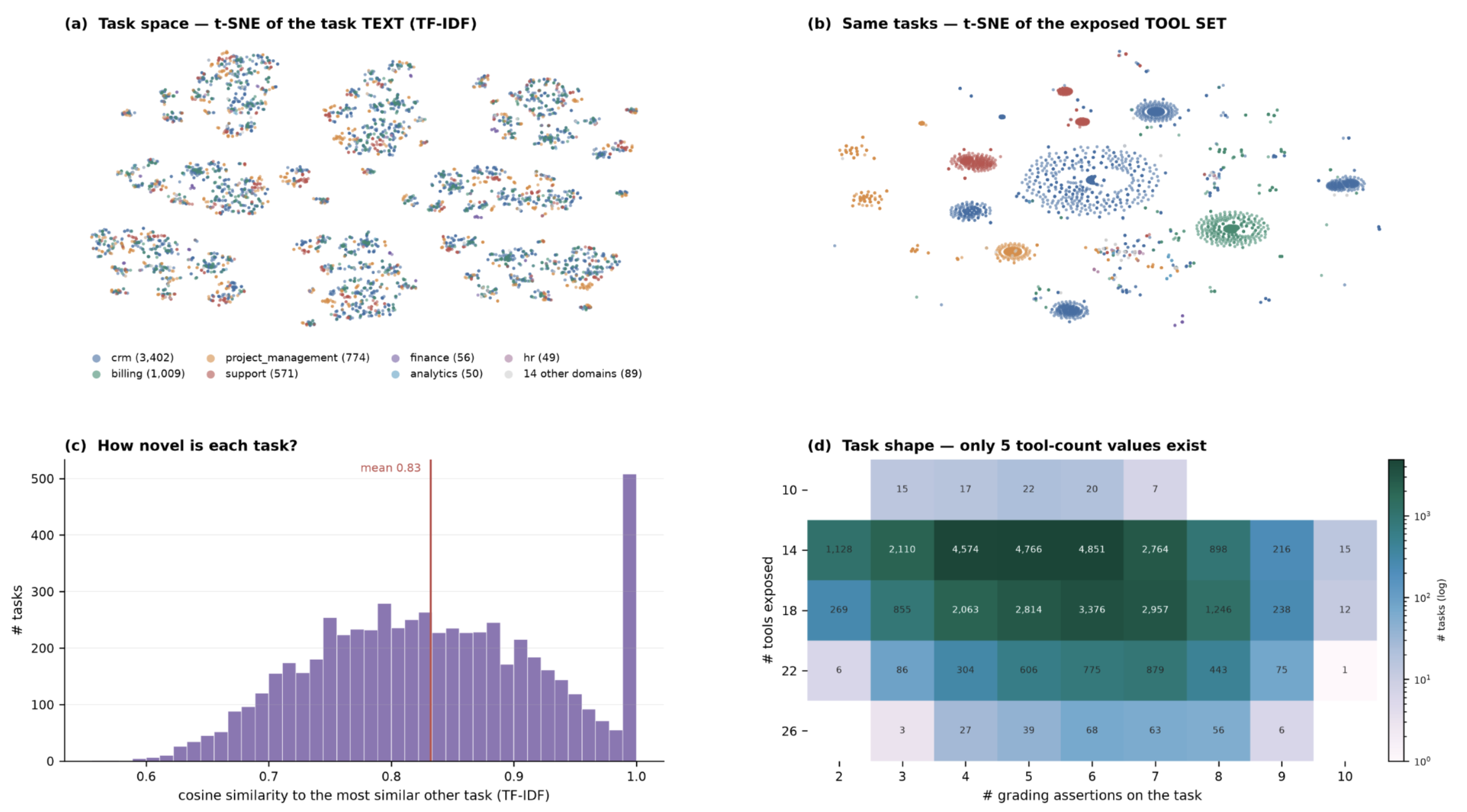}
    \caption{
    Structural analysis of the synthesized task space.
(a) A t-SNE projection of task descriptions based on TF-IDF features shows substantial textual clustering, but the clusters do not correspond to meaningful task metadata such as domain, persona, or task shape.
(b) In contrast, projecting the same tasks using their exposed tool sets reveals clear domain-specific structure, indicating that the primary source of task variation lies in the executable tool surface rather than in task language.
(c) Distribution of nearest-neighbor cosine similarity in the TF-IDF space, showing that although the overall task space is broad, most tasks still have a highly similar counterpart.
(d) The number of exposed tools takes only five values, $\{10,14,18,22,26\}$, revealing a discrete task-shape lattice induced by the core tool set and the number of additional resources.
    }
    \label{fig:task_diversity}
    \vspace{-0.2cm}
\end{figure}

\paragraph{Task diversity is primarily expressed through tool structure.}
Figure~\ref{fig:task_diversity} analyzes the structural diversity of
the synthesized task corpus from four complementary views.
We distinguish between the linguistic surface of a task, the executable
tool surface exposed to the agent, the local textual similarity between
tasks, and the combinatorial structure of task size.

\subsection{Task Diversity}
Figure~\ref{fig:task_diversity}(a) shows a t-SNE projection of task descriptions using TF-IDF features.
Although the projection contains several visually distinct blobs, these clusters do not correspond to meaningful task metadata.
We applied $k$-means clustering with $k=9$ to the projected task space and compared the composition of each cluster across primary domain, \texttt{persona\_role}, the AF flag, number of exposed tools, and source shard.
The resulting clusters exhibit nearly identical compositions across these metadata dimensions: CRM accounts for approximately 53-61\% of each cluster, Account Manager is the dominant persona in approximately 53-63\%, the AF flag is present in approximately 58-61\%, and tasks with 14 exposed tools account for approximately 52-61\%.
Thus, the apparent textual blobs are not semantic partitions of the task corpus.
This suggests that textual embedding distance is a poor proxy for task-level diversity in this corpus: the task descriptions share a common template and vary primarily through their underlying executable structure.

To examine whether the same tasks exhibit structure in their executable interfaces, we represent each task by a binary vector over the 750 available tools and project the resulting vectors using SVD followed by cosine t-SNE.
In contrast to the textual representation, the same set of tasks forms clearly separated, domain-specific clusters.
In particular, CRM, billing, project management, and support tasks form distinct dense regions in the projected tool space.
Because the tasks and samples are identical between Figures~\ref{fig:task_diversity}(a) and~\ref{fig:task_diversity}(b), the difference is attributable to the representation rather than to a change in the underlying data.
Together, the two views indicate that the dominant source of meaningful variation in the corpus is the tool surface exposed to the agent, rather than lexical variation in the task description.

Figure~\ref{fig:task_diversity}(c) further characterizes the local structure of the task corpus by measuring each task's cosine similarity to its nearest neighbor in TF-IDF space.
The corpus has a low mean pairwise similarity of 0.13, indicating a broad overall textual space, while the nearest-neighbor distribution is substantially more concentrated, with a median similarity of 0.83.
Moreover, 8.3\% of tasks have a nearest neighbor with cosine similarity above 0.99.
Thus, the corpus simultaneously contains broad global variation and a non-negligible population of near-twin tasks.

Importantly, only 17\% of these near-twin pairs originate from the same environment as sibling seeds.
If the observed redundancy were primarily caused by the ten-seed replication within each environment, approximately 78\% of such pairs would be expected to be within-environment siblings.
The observed proportion therefore suggests that the local redundancy is not explained by seed replication alone, but instead arises substantially from shared task templates across different environments.

Finally, Figure~\ref{fig:task_diversity}(d) examines the combinatorial structure of task size.
The number of exposed tools takes only five distinct values,
$\{10,14,18,22,26\}$.
This structure follows directly from a fixed core of ten tools, consisting of employee search/list/get/update, email search/list, send-email, and Slack list-channels/list-messages/post-message operations.
Each additional resource contributes exactly four tools corresponding to search, list, get, and update operations.
Consequently, task size is not continuously distributed: it is determined largely by a single discrete variable, namely the number of additional resources.
The resulting task space therefore forms a low-dimensional lattice of task shapes rather than an unconstrained continuum.

\section{Hyperparameter Details}
\label{app:hyperparameter}

We provide the detailed training configuration used for policy optimization with \textsc{AgentMercury} to facilitate reproducibility.
Unless otherwise specified, the configuration below is used for the Qwen3.5-4B experiments reported in the main paper.
We organize the configuration into four aspects: model and memory configuration, rollout and asynchronous execution, executable-environment interaction, and policy optimization and reward computation.

\paragraph{Model and Memory Configuration.}
We use tensor parallelism across two GPUs, while pipeline, context, and expert parallelism are disabled.
Sequence parallelism is enabled to reduce the memory footprint of long-context training.
We use full recomputation with one layer of uniform recomputation and the FlashAttention backend.
The attention softmax is computed in FP32 and dropout is disabled.
The maximum number of tokens per GPU is set to $24{,}576$.

\begin{table}[t]
    \centering
    \small
    \caption{
    Model-parallelism and memory configuration used for
    Qwen3.5-4B policy optimization.
    }
    \label{tab:hyper_model}
    \begin{tabular}{@{}ll@{}}
        \toprule
        \textbf{Parameter} & \textbf{Value} \\
        \midrule
        Tensor parallelism & 2 \\
        Pipeline / context / expert parallelism & 1 / 1 / 1 \\
        Sequence parallelism & Enabled \\
        Recomputation & Full, uniform, 1 layer \\
        Attention backend & FlashAttention \\
        Attention softmax precision & FP32 \\
        Dropout & 0 \\
        Max tokens per GPU & 24,576 \\
        \bottomrule
    \end{tabular}
\end{table}

\paragraph{Rollout and Asynchronous Execution.}
For agent rollouts, we allow responses of up to $16{,}384$ tokens with a SGLang context length of $24{,}576$.
Rollouts are sampled at temperature $1.0$ and terminate at the end-of-turn token \texttt{<|im\_end|>}.
We enable partial rollouts and use the Qwen3 and Qwen3-Coder parsers for reasoning and tool calls. The SGLang memory fraction is set to $0.8$, with at most 64 concurrently running requests.

We use fully asynchronous policy optimization to overlap actor updates and environment rollouts.
Two GPUs are assigned to the actor and six GPUs to rollout generation, for a total of eight GPUs on a single node.
The maximum allowed policy staleness is four updates, while the rollout in-flight cap is 64.
We require a minimum fresh-token ratio of $0.75$ and update rollout workers after every optimization step.

\begin{table}[t]
    \centering
    \small
    \caption{
    Rollout and fully-asynchronous execution configuration.
    }
    \label{tab:hyper_rollout}
    \begin{tabular}{@{}ll@{}}
        \toprule
        \textbf{Parameter} & \textbf{Value} \\
        \midrule
        Max response length & 16,384 \\
        SGLang context length & 24,576 \\
        Rollout temperature & 1.0 \\
        End-of-turn token & \texttt{<|im\_end|>} \\
        Partial rollout & Enabled \\
        Reasoning / tool parser & Qwen3 / Qwen3-Coder \\
        SGLang memory fraction & 0.8 \\
        Max running requests & 64 \\
        Fully asynchronous training & Enabled \\
        Actor / rollout GPUs & 2 / 6 \\
        Max staleness & 4 \\
        Rollout in-flight cap & 64 \\
        Minimum fresh-token ratio & 0.75 \\
        Weight update interval & 1 \\
        Fault tolerance & Enabled \\
        \bottomrule
    \end{tabular}
\end{table}

\paragraph{Executable Environment Configuration.}
Each rollout interacts with an executable MCP-based environment.
We allow up to 20 tool-use turns per episode and up to 8,192 generated tokens per turn.
The environment runs locally in-process, avoiding network overhead during training.
The calendar tool is used as the probe tool with a creation throttle of eight.
Environment readiness and reset timeouts are set to 600 and 180 seconds, respectively.

\begin{table}[t]
    \centering
    \small
    \caption{
    Configuration of the executable MCP environments used during policy
    optimization.
    }
    \label{tab:hyper_environment}
    \begin{tabular}{@{}ll@{}}
        \toprule
        \textbf{Parameter} & \textbf{Value} \\
        \midrule
        Maximum MCP turns & 20 \\
        Maximum generation tokens / turn & 8,192 \\
        Arena backend & Local (in-process) \\
        Probe tool & Calendar \\
        Environment creation throttle & 8 \\
        Ready timeout & 600 s \\
        Reset timeout & 180 s \\
        \bottomrule
    \end{tabular}
\end{table}

\paragraph{Reward Computation.}
The reward is computed by the \textsc{AgentMercury} reward implementation from the final environment state and the agent trajectory.
We use the fraction aggregator and enable the behavior penalty.
In particular, the reward implementation includes safeguards against undesirable generation behavior, including phrase repetition, post-answer continuation, and degeneracy-aware truncation handling.
The reward judge is DeepSeek-V4-Flash with temperature $0.0$ and a maximum generation length of 4,000 tokens.
During cold start, truncated samples are retained rather than discarded so that the truncation penalty remains part of the learning signal.

\begin{table}[t]
    \centering
    \small
    \caption{
    Reward and verification configuration used during RL training.
    }
    \label{tab:hyper_reward}
    \begin{tabular}{@{}ll@{}}
        \toprule
        \textbf{Parameter} & \textbf{Value} \\
        \midrule
        Reward implementation
            & \textsc{AgentMercury} \texttt{rl\_reward.py} \\
        Reward aggregator & Fraction \\
        Behavior penalty & Enabled \\
        Reward judge & DeepSeek-V4-Flash \\
        Judge temperature & 0.0 \\
        Judge max tokens & 4,000 \\
        Truncated-sample handling & Retained during cold start \\
        \bottomrule
    \end{tabular}
\end{table}

\paragraph{GRPO and Optimization.}
For the main Qwen3.5-4B experiments, we use GRPO with eight samples per prompt.
The global batch size is 128 and the rollout batch size is 16.
We use a decoupled-PPO surrogate with asymmetric clipping parameters of $0.2$ and $0.28$ following Dr.GRPO~\citep{liu2025understanding}.
GRPO standard-deviation normalization and the entropy regularizer are disabled.
We additionally filter groups whose rewards have zero standard deviation, as these groups provide no relative learning signal.
The policy is optimized with Adam using $\beta_1=0.9$ and $\beta_2=0.98$.
The learning rate is fixed at $1\times10^{-6}$ without warmup, and weight decay is set to zero.
Per-token loss computation is enabled.

\begin{table}[t]
    \centering
    \small
    \caption{
    GRPO and optimizer configuration for Qwen3.5-4B.
    }
    \label{tab:hyper_grpo}
    \begin{tabular}{@{}ll@{}}
        \toprule
        \textbf{Parameter} & \textbf{Value} \\
        \midrule
        Advantage estimator & GRPO \\
        Surrogate & Decoupled-PPO (\texttt{icepop}) \\
        $\epsilon$-clip / $\epsilon$-clip-high & 0.20 / 0.28 \\
        Entropy coefficient & 0 \\
        GRPO std normalization & Disabled \\
        Per-token loss & Enabled \\
        Global batch size & 128 \\
        Rollout batch size & 16 \\
        Samples per prompt & 8 \\
        Dynamic sampling filter
            & Zero-std reward groups dropped \\
        \midrule
        Optimizer & Adam \\
        $\beta_1$ / $\beta_2$ & 0.9 / 0.98 \\
        Learning rate & $1\times10^{-6}$ \\
        Learning-rate schedule & Constant \\
        Warmup & 0 \\
        Weight decay & 0 \\
        Gradient accumulation & FP32 \\
        \bottomrule
    \end{tabular}
\end{table}

\paragraph{Training Corpus.}
The Qwen3.5-4B policy is initialized from the Qwen3.5-4B base checkpoint and trained on the synthesized \textsc{AgentMercury} investigation corpus.
The corpus contains 43,300 executable tasks generated from the synthesized environment library, with ten seeded tasks per environment configuration.
The same corpus is sampled throughout RL training rather than being constructed specifically for any of the downstream benchmark evaluations.

\begin{table}[t]
    \centering
    \small
    \caption{
    Model and corpus configuration for the Qwen3.5-4B RL experiment.
    }
    \label{tab:hyper_data}
    \begin{tabular}{@{}ll@{}}
        \toprule
        \textbf{Parameter} & \textbf{Value} \\
        \midrule
        Base model & Qwen3.5-4B \\
        Training corpus & \textsc{AgentMercury} investigation corpus \\
        Number of tasks & 43,300 \\
        Seeds per environment & 10 \\
        \bottomrule
    \end{tabular}
\end{table}

Overall, these settings are designed to make long-horizon interaction with executable environments practical while preserving a sufficiently rich learning signal for policy optimization.
In particular, the combination of asynchronous rollouts, partial generation, dynamic reward filtering, and executable environment verification allows the policy to be trained directly on multi-turn tool-use trajectories without reducing the environments to static instruction-following examples.

\paragraph{SAO Configuration.}
We additionally investigate SAO as an alternative policy optimization procedure.
We use the same synthesized environments and broadly comparable rollout infrastructure to isolate the effect of the optimization procedure.
However, we find that the Qwen3.5-4B policy provides a substantially weaker learning signal under SAO: the training dynamics exhibit under-representation of useful updates compared with GRPO.
We therefore do not use the 4B SAO run as a primary comparison in the main benchmark table.
The corresponding training dynamics are reported in Appendix~\ref{fig:35b_training_log_sao} and discussed further in Section~\ref{sec:discussion}.

\section{Benchmark Details}
\label{app:benchmark}
\paragraph{Common Evaluation Setup.}
Unless otherwise specified, all models are served using SGLang with the \texttt{slimerl/slime} Docker environment and exposed through an OpenAI-compatible endpoint.
We use the Qwen3 reasoning parser and Qwen3-Coder tool-call parser with \texttt{--trust-remote-code} and a static memory fraction of 0.85 (or 0.90 for the 35B model).
The native maximum position length is 262K tokens without additional RoPE scaling~\citep{}.
For the 4B model, we use a 32K context window for mathematical, coding, and general agentic benchmarks, and a 128K context window for \textsc{EnterpriseOps-Gym}.
The 35B-A3B model is evaluated with a 128K context window across all benchmarks.
Tensor parallelism is set to 1 for the 4B model at 32K context, 2 for the 4B model at 128K context, and 4 for the 35B-A3B model.

Each benchmark is independently evaluated three times, and we report the mean and standard deviation across runs.
These independent repetitions are performed in addition to
benchmark-specific sampling or trial repetitions, such as 10 samples per
problem for AIME, 5 for GPQA-Diamond, 4 for HMMT, 4 trials for tau-3 benchmark, and 3 runs for \textsc{EnterpriseOps-Gym}.
Unless otherwise specified, sampling uses temperature 1.0, with \texttt{top\_p}=0.95 for interactive agent benchmarks.

We use programmatic evaluation whenever the benchmark provides a deterministic verifier.
Specifically, AIME, HMMT, LiveCodeBench, SciCode, BFCL, and GPQA-Diamond
are evaluated without an LLM judge using exact-match, code-execution, AST/executable-call, or multiple-choice verification, as appropriate.
LLM-based judging is used only for interactive conversational evaluation in tau-3 benchmark, where DeepSeek-V4-Flash is used through DashScope.
EnterpriseOps-Gym uses its deterministic oracle verifier.
For GPQA-Diamond, generations with a score below 0.30, corresponding to truncated or effectively empty responses, are discarded and re-executed according to the evaluation guard.

For model-scale comparisons, the 4B base and iteration-149 checkpoints use the 32K configuration for core mathematical, coding, and agentic benchmarks and the 128K configuration for EnterpriseOps-Gym.
The 35B-A3B model uses the 128K configuration for all benchmarks.
For the 35B-A3B evaluation, two independent endpoints are deployed across GPU groups to parallelize long-running evaluations; for example, the 990 GPQA-Diamond generations are distributed across the two endpoints.
All evaluation runs use the same benchmark versions, evaluation identifiers, and scoring procedures to ensure comparability across model sizes.

\paragraph{AIME 2026.}
AIME 2026~\citep{aime26} consists of 30 competition-level mathematical problems.
We sample each problem 10 times, resulting in 300 attempts per evaluation run, and compute the average accuracy (\textit{avg@10}) using exact matching against the integer-valued ground-truth answers.
Evaluation is fully programmatic and does not involve an LLM judge.
The benchmark primarily evaluates multi-step mathematical reasoning under competition-style problem solving.

\paragraph{HMMT February 2026.}
HMMT February 2026~\citep{dekoninck2026matharena} is a mathematics competition benchmark based on the February 2026 Harvard-MIT Mathematics Tournament.
We evaluate approximately 30 problems with four samples per problem (\textit{num\_execute}=4), and report the mean exact-match accuracy over final answers.
As a recently administered competition, the benchmark provides an additional evaluation setting with reduced exposure to training-time contamination.
The problems emphasize challenging mathematical reasoning, with final answers that can be programmatically verified.

\paragraph{LiveCodeBench v5--v6.}
We use the LiveCodeBench v5--v6 split~\citep{jain2025livecodebench}, which contains only problems newly introduced between LiveCodeBench versions 5 and 6.
This setting is designed to reduce contamination from previously released competitive-programming problems.
Generated programs are executed against hidden tests and evaluated by pass/fail correctness, with \textit{pass@1} reported as the primary metric.
The benchmark evaluates the complete programming pipeline from problem specification understanding and algorithmic reasoning to executable code generation.

\paragraph{SciCode.}
SciCode~\citep{tian2024scicode} evaluates scientific computing ability across domains including physics, chemistry, biology, and materials science.
The main test split contains 65 problems, each decomposed into dependent subproblems that collectively require domain-specific reasoning and code generation.
Solutions are evaluated by executing the generated code against reference implementations and associated H5 datasets, with both subproblem-level and main-problem pass rates used for evaluation.
The benchmark therefore emphasizes domain-grounded computational reasoning rather than generic code generation.

\paragraph{Tau-3.}
Tau-3 benchmark~\citep{taubench, tau2bench, shi2026tau} evaluates interactive tool-use agents across airline, retail, and telecom customer-service domains.
An agent interacts with a simulated user through multi-turn conversations and executable tool calls to accomplish task-specific goals.
We use four trials per task (\textit{num\_trials}=4) and report the average \textit{pass\textsuperscript{1}} reward.
Task success is evaluated through state read-back together with an LLM-based natural-language assertion judge using DeepSeek-V4-Flash~\citep{deepseekai2026deepseekv4}.
We focus on the three supported domains and omit the banking-knowledge retrieval setting because it requires a 128K context window that is not available under our corresponding evaluation configuration.
This benchmark isolates interactive agentic capabilities involving multi-turn reasoning and tool use rather than standalone knowledge recall.

\paragraph{BFCL.}
The Berkeley Function-Calling Leaderboard (BFCL)~\citep{patil2025berkeley} evaluates whether an agent
can produce correct tool calls, including the appropriate functions, arguments, and argument types.
We use its programmatic evaluation based on AST matching and executable function-call verification, and report overall accuracy.
Unlike long-horizon interactive benchmarks, BFCL primarily measures the structural precision of tool calling and function-argument generation.

\paragraph{GPQA-Diamond.}
GPQA-Diamond~\citep{rein2023gpqa} is the expert-validated ``Diamond'' subset of GPQA, consisting of 198 graduate-level multiple-choice questions spanning physics, chemistry,
and biology.
We evaluate each question five times, resulting in 990 attempts per evaluation run, and report mean multiple-choice accuracy.
The benchmark emphasizes knowledge-intensive reasoning on questions designed to be difficult to answer through straightforward retrieval or search.

\paragraph{\textsc{EnterpriseOps-Gym}.}
\textsc{EnterpriseOps-Gym}~\citep{malay2026enterpriseops} is an enterprise-agent benchmark covering eight business domains: teams, customer success management (CSM), email, IT service management (ITSM), calendar, drive, hybrid, and human resources (HR).
We evaluate agents using the oracle setting, in which a task is considered successful only when all task-specific verifiers are satisfied.
Agents interact with isolated task databases through MCP-based tool servers and execute multi-step workflows using a ReAct-style interaction loop.
We run each task three times (\textit{num\_runs}=3) and report the resulting \textit{pass@1} accuracy.
The benchmark provides a particularly direct evaluation of business-oriented agentic behavior, including multi-step tool use, state manipulation, and, for hybrid tasks, coordination across multiple service domains.

\section{Training Log}
\begin{figure}[t]
    \centering
    \includegraphics[width=\columnwidth]{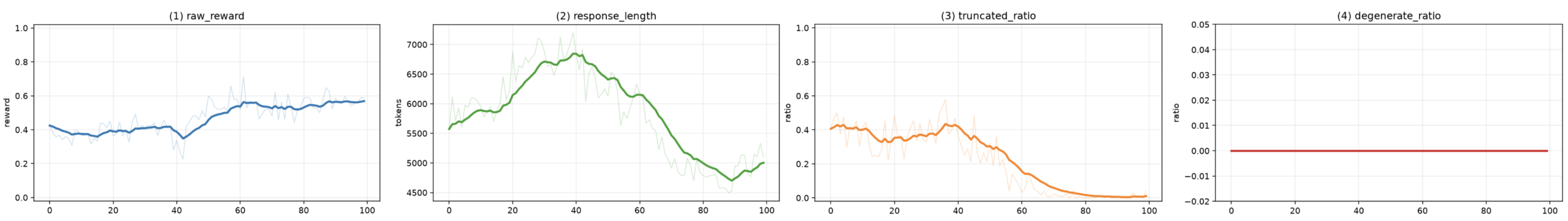}
    \caption{
    Training dynamics of Qwen3.5-35B-A3B with GRPO and
    \textsc{AgentMercury}.
    We report the raw reward, response length, truncated-response ratio, and degenerate-response ratio throughout training.
    The increasing reward is accompanied by a steady reduction in truncation, while degenerate responses remain negligible.
    }
    \label{fig:35b_training_log}
    \vspace{-0.2cm}
\end{figure}

\begin{figure}[t]
    \centering
    \includegraphics[width=\columnwidth]{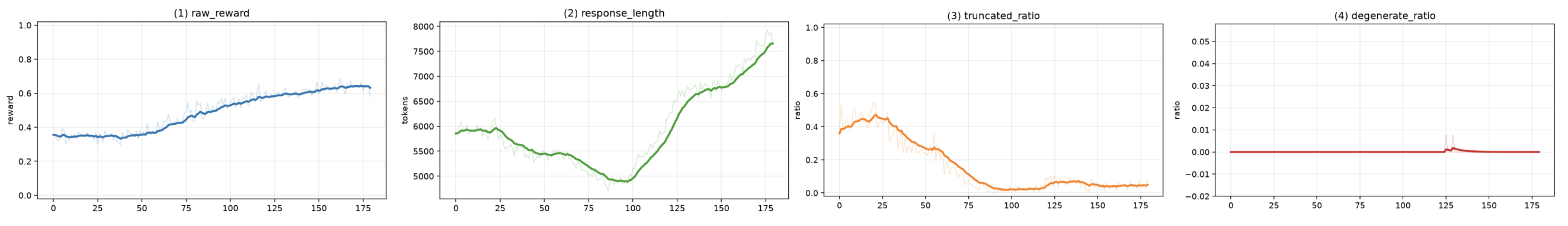}
    \caption{
    Training dynamics of Qwen3.5-35B-A3B with SAO and
    \textsc{AgentMercury}.
    We report the raw reward, response length, truncated-response ratio, and degenerate-response ratio throughout training.
    SAO exhibits a sustained reward increase together with a substantial reduction in truncation, while degenerate responses remain close to zero.
    }
    \label{fig:35b_training_log_sao}
    \vspace{-0.2cm}
\end{figure}

We provide additional training diagnostics for the Qwen3.5-35B-A3B experiments in Figures~\ref{fig:35b_training_log} and~\ref{fig:35b_training_log_sao}.
We monitor four quantities throughout training: raw reward, response length, truncated-response ratio, and degenerate-response ratio.
These diagnostics are particularly useful for checking whether the improvement in policy reward is accompanied by undesirable generation behavior such as excessive response truncation or degenerate outputs.

For GRPO, the raw reward initially remains around $0.4$ before increasing steadily during the latter half of training and reaching approximately $0.55$-$0.60$ toward the end of the run (Figure~\ref{fig:35b_training_log}).
At the same time, the truncated-response ratio decreases from roughly $0.35$ at the beginning of training to nearly zero toward the end.
The response length initially increases, reaching its maximum around the middle of training, and subsequently decreases as training progresses.
Despite these changes in response length, the reward continues to improve, suggesting that the later-stage reward improvement is not driven simply by generating increasingly long responses.
The degenerate-response ratio remains effectively zero throughout the run.

The SAO run exhibits a similar overall pattern, while covering a longer training trajectory (Figure~\ref{fig:35b_training_log_sao}).
The raw reward increases from approximately $0.35$ at initialization to above $0.6$ near the end of training.
The truncated-response ratio decreases substantially from approximately $0.4$ at the beginning to around $0.05$ in the later stages, although small fluctuations remain.
Response length first decreases during the early and middle stages of training, reaches a minimum around the middle of the run, and then increases substantially during the later stages. 
The degenerate-response ratio remains close to zero for nearly the entire run, with only a small transient spike in the later stages.

Overall, both optimization methods show that the policy can improve its reward while simultaneously reducing severe generation failures.
In particular, the consistent decrease in truncated responses indicates that the synthesized environments provide a sufficiently structured learning signal for the policy to learn executable, complete interaction trajectories rather than exploiting the reward through increasingly malformed outputs.
The near-zero degenerate-response ratio further suggests that the observed reward improvement is not accompanied by a collapse into degenerate generation behavior.

\section{Additional Details on Environment Authoring}
\label{sec:app_authoring}

This appendix provides additional details on the environment-authoring experiment described in Section~\ref{sec:exp_authoring}.
Our goal is to distinguish between two questions: whether a capable model can produce a structurally valid executable world from a high-level scenario, and whether the construction procedure exposed by \textsc{AgentMercury} can be learned through supervision.
We therefore evaluate both zero-shot authoring and authoring after fine-tuning on construction traces.

\paragraph{Authoring Protocol.}
We construct a held-out set of 30 synthetic business briefs sampled from the country--industry distribution of the environment library. Each brief describes a high-level business scenario while withholding the corresponding environment specification, including its services, state tables, tools, and cross-service invariants.
The task given to the model is to transform the brief into a complete executable environment specification.

We evaluate two prompting conditions.
In the \emph{zero-shot} condition, the model receives only the high-level business brief.
In the \emph{recipe} condition, the model additionally receives an invariant digest describing the structural requirements together with a trimmed exemplar of a previously constructed environment.
The latter condition tests whether explicitly exposing the construction procedure can improve authoring reliability without requiring parameter updates.

The generated environments are evaluated using the same executable construction oracle underlying \textsc{AgentMercury}.
In particular, we use 12 structural validators covering the consistency and executability of the generated world.
A generation is counted as an oracle success only when all 12 validators pass.
Thus, the metric is deliberately stricter than textual or semantic similarity: a model must produce a world that can actually be executed while satisfying the required structural constraints.

\paragraph{Statistical Analysis.}
Because the same 30 briefs are evaluated under both prompting conditions, we treat the comparison as paired rather than independent.
Across the five off-the-shelf API models, zero-shot authoring achieves a mean oracle-pass rate of $80.7\%$, compared with $78.0\%$ under recipe conditioning.
The paired comparison does not reveal a systematic advantage for providing the recipe; the exact McNemar test gives $p \geq 0.375$ for every individual model.
Differences between the two conditions are therefore better interpreted as prompt sensitivity than as evidence that explicit construction instructions improve authoring reliability.

For the fine-tuned Qwen3.5-35B-A3B model, the zero-shot oracle-pass rate increases from $3.3\%$ to $83.3\%$. The corresponding Wilson $95\%$ confidence intervals are $[0.6,16.7]$ for the base model and $[66.4,92.7]$ after fine-tuning.
A Fisher exact test gives $p=1.2\times10^{-10}$, providing strong evidence that the improvement is not explained by sampling variation over the 30 held-out briefs.

\paragraph{Failure Modes of Zero-shot Authoring.}
Although the strongest API models frequently produce valid executable worlds, the remaining failures are not uniformly distributed across the validators. The dominant failure mode involves \emph{cross-service invariants}: the generated environment specifies a constraint whose trigger and target should belong to different services, but instead places both sides within a single service.
This failure is particularly pronounced for GPT-5.4, where 10 of the 30 held-out briefs exhibit such a violation.

This observation highlights the importance of executable validation.
A generated specification can be syntactically complete and superficially plausible while still failing to represent the intended interaction structure between services.
Such errors are difficult to identify reliably through surface-level inspection alone, but are directly exposed by the executable oracle.
The construction oracle therefore serves not only as an evaluation metric but also as a mechanism for identifying structural weaknesses in environment authoring.

We additionally observe formatting and parsing failures for some open models under recipe conditioning.
In particular, providing a longer construction recipe can introduce additional formatting requirements that are themselves a source of failure.
This explains why recipe conditioning does not monotonically improve authoring performance, despite providing the model with more explicit information about the desired environment structure.

\paragraph{Construction Traces as Supervision.}
To test whether the authoring process itself is learnable, we fine-tune Qwen3.5-35B-A3B on 29,823 samples derived from the construction traces of \textsc{AgentMercury}.
Rather than training exclusively on final brief-to-world pairs, the supervision covers several stages of the construction process:

\begin{itemize}
    \item \textbf{Brief-to-world generation}, mapping a high-level business
    scenario to a complete executable world;
    \item \textbf{Intermediate-stage completion}, requiring the model to
    complete partially constructed environment components;
    \item \textbf{Validator-guided corruption repair}, where the model
    receives an invalid construction and validator feedback and must repair
    the corresponding structural error;
    \item \textbf{Intent-to-diff prediction}, mapping a high-level intent to
    the corresponding structural modification of an existing environment.
\end{itemize}

This mixture exposes the model to both forward construction and error-correction trajectories.
In contrast to ordinary supervised specification generation, the model therefore observes intermediate construction states and the consequences of violating environment invariants.

Before fine-tuning, Qwen3.5-35B-A3B successfully passes all 12 validators on only $3.3\%$ of held-out briefs.
The base model also frequently truncates or breaks the required output format.
After fine-tuning, the oracle-pass rate reaches $83.3\%$, while the average number of passed validators increases to $11.5$ out of 12. Truncation and formatting failures are nearly eliminated.
Thus, the improvement is not merely due to producing longer or more well-formed specifications; the trained model learns to satisfy the structural constraints required for executable worlds.

\paragraph{Prompting Versus Internalized Construction.}
An additional analysis reveals an interesting interaction between recipe conditioning and learned construction behavior.
For the base Qwen3.5-35B-A3B model, providing the construction recipe increases the oracle-pass rate from $3.3\%$ to $20.0\%$.
The same recipe, however, reduces the performance of the fine-tuned model from $83.3\%$ to $10.0\%$.

The failure of the fine-tuned model is highly concentrated: 27 of the 30 recipe-conditioned generations fail the cross-service validation check.
This asymmetric behavior suggests that prompting and parameter-level learning provide qualitatively different forms of procedural knowledge.
For the base model, the recipe supplies information that is otherwise absent from its generation policy.
After fine-tuning, however, the construction procedure is already encoded in the model parameters, and the additional recipe can interfere with the learned generation format and structural assumptions.

This result also provides evidence against interpreting the observed improvement as simple instruction following.
If the fine-tuned model merely benefited from additional textual instructions, recipe conditioning should continue to improve performance after training.
Instead, the learned policy performs best when it is given the original high-level brief without the additional construction recipe.

The authoring experiment demonstrates two complementary properties of \textsc{AgentMercury}.
First, executable-world construction is already within the capability range of sufficiently strong general-purpose models, although cross-service structural constraints remain a significant source of failure.
Second, the construction process can be substantially improved through training on construction traces.
Fine-tuning converts Qwen3.5-35B-A3B from a model that passes the complete construction oracle on only 3.3\% of held-out briefs into one achieving 83.3\% oracle success, with nearly complete validator coverage.

Together with the main results in Section~\ref{sec:exp_authoring}, these findings suggest that \textsc{AgentMercury} exposes two learnable interfaces:
the synthesized executable world can provide a training substrate for an agent policy, while the construction traces can provide supervision for learning how to create such worlds.

\end{document}